\pdfoutput=1
\documentclass[review]{elsarticle}
\usepackage{subfigure}
\usepackage{longtable}
\usepackage{algorithm, algorithmic}
\usepackage{bm}
\usepackage{booktabs}
\usepackage{multirow}
\usepackage{geometry}
\usepackage{amsfonts,amssymb}

\usepackage{lineno,hyperref}
\modulolinenumbers[5]

\journal{arXiv}

\begin{document}

\begin{frontmatter}

\title{Universal Transformer Hawkes Process with Adaptive Recursive Iteration}


\author[mymainaddress]{Lu-ning Zhang}
\ead{luning_zhang@foxmail.com}

\author[mymainaddress]{Jian-wei Liu\corref{mycorrespondingauthor}}
\cortext[mycorrespondingauthor]{Corresponding author}
\ead{liujw@cup.edu.cn}

\author[mymainaddress]{Zhi-yan Song}

\author[mymainaddress]{Xin Zuo}

\address[mymainaddress]{Department of Automation, College of Information Science and Engineering,
China University of Petroleum , Beijing, Beijing, China}

\begin{abstract}
Asynchronous events sequences are widely distributed in the natural world and human activities, such as earthquakes records, users’ activities in social media and so on. How to distill the information from these seemingly disorganized data is a persistent topic that researchers focus on. The one of the most useful model is the point process model, and on the basis, the researchers obtain many noticeable results. Moreover, in recent years, point process models on the foundation of neural networks, especially recurrent neural networks (RNN) are proposed and compare with the traditional models, their performance are greatly improved. Enlighten by transformer model, which can learning sequence data efficiently without recurrent and convolutional structure, transformer Hawkes process is come out, and achieves state-of-the-art performance. However, there is some research proving that the re-introduction of recursive calculations in transformer can further improve transformer’s performance. Thus, we come out with a new kind of transformer Hawkes process model, universal transformer Hawkes process (UTHP), which contains both recursive mechanism and self-attention mechanism, and to improve the local perception ability of the model, we also introduce convolutional neural network (CNN) in the position-wise-feed-forward part. We conduct experiments on several datasets to validate the effectiveness of UTHP and explore the changes after the introduction of the recursive mechanism. These experiments on multiple datasets demonstrate that the performance of our proposed new model has a certain improvement compared with the previous state-of-the-art models.
\end{abstract}


\end{frontmatter}

\section{Introduction}

In this era of informatization and data in trajectories, nature and human activities will be recorded as a large amount of asynchronous sequential data, for instance, the record of occurrence of earthquakes and aftershocks \cite{1ogata1998space}, transaction history of financial markets \cite{2bacry2015hawkes}, electronic health record \cite{3wang2018supervised}, equipment failure history \cite{4zhang2020survival} and user behavior in social networks \cite{5zhou2013learninga,6zhao2015seismic}, such as Weibo, Twitter, Facebook, etc. all are asynchronous events sequences data.

These asynchronous sequences contain valuable knowledge and information, researchers utilize a variety of methods to better mine these knowledge and information from the data. Among a diversity of methods, point process is one of the most widely used methods in this field, and in point process models, Hawkes process \cite{7hawkes1971spectra} is one of the most commonly used model. The self-exciting progress in Hawkes process fits the interaction between events to some extent, thus the application of Hawkes process achieves certain results in sequence analysis.

For example, Zhou et al. present alternating direction method of multipliers (ADMM)-based algorithm to learn Hawkes process to discover the hidden network of social influence \cite{7hawkes1971spectra}, Xu et al. make use of non-parametric Hawkes process to reveal the Granger causality between the users activity on watching internet protocol television (IPTV) \cite{8xu2016learning}. Hansen et al. demonstrate the great expressive potential of Hawkes process in neuroscience with least absolute shrinkage and selection operator (LASSO) method \cite{9hansen2015lasso}. Reynaud-Bouret and Schbath provide a new way to detect the favored or avoided distances between genomic events along deoxyribonucleic acid (DNA) sequences \cite{10reynaud2010adaptive}. Zhang et al. \cite{4zhang2020survival} modify the traditional Hawkes process, introducing the time-dependent background intensity to Hawkes process to analyze the background probability of failure and relationship between failures in compressor station.

However, traditional Hawkes process only considers the positive superposition of effect of historical events, which severely constrains the fitting ability of this type of model. Meanwhile, the lack of nonlinear operations in traditional Hawkes process also sets an upper limit for Hawkes’ expressive ability. Thus, in recent years, due to the strong fitting ability of neural networks and especially the sequence modeling ability of RNN, the research direction in this field is transferred to the neural point process model. For instance, Du et al. \cite{11du2016recurrent} embed the information of sequence data (including time-stamp and event type) into RNN, and come up with the recurrent marked temporal point processes (RMTPP) to model the conditional intensity function by nonlinear dependency of the history. And similar to RMTPP, Mei et al. \cite{12mei2017neural} propose the continuous-time LSTM (Long Short-Term Memory) to model the conditional intensity of point process, which is called as neural Hawkes process, and in this continuous neural Hawkes process, the influence of the previous event decays with time continuously. Xiao et al. use two RNN to model the conditional intensity function, one is used to process time stamp information, and the other is used to process historical event information \cite{13xiao2017modeling}.

Inevitably, these RNN based-models also inherit the drawbacks of RNN, for instance, it may take a long time for the patient to develop symptoms due to certain sequel, which has obvious long-term characteristics, such as diabetes, cancer and other chronic diseases, while these RNN-based models are hard to reveal the long-term dependency between the distant events in the sequences \cite{14bengio1994learning}. The ideal point process model should be able to solve these problems. Moreover, in the training of RNN-based models, such problems as vanishing and exploding Gradients \cite{15pascanu2013difficulty} often occur, and then affect the performance of the model.

It’s worth noting that in traditional sequence learning problem, such as machine translation \cite{16raganato2018analysis} and speech recognition \cite{17dong2018speech}, transformer model \cite{18vaswani2017attention} based on self-attention \cite{19DBLP:journals/corr/BahdanauCB14} mechanism achieves distinct performance improvement without application of CNN and RNN, meanwhile, the transformer structure free-from recurrent modules makes the model have higher computational efficiency. These achievements give a new insight on the development of sequential data learning. On account of this fact, Zhang et al. present self-attention Hawkes process \cite{20zhang2020self}, furthermore, Zuo et al. \cite{21DBLP:conf/icml/ZuoJLZZ20} propose transformer Hawkes process based on the attention mechanism and encoder structure in transformer. This model utilizes pure transformer structure without using RNN and CNN, and achieves state-of-the-art performance, but there is still much room for improvement in the transformer models, for instance, transformer simply stack the encoder layer to learn sequence data, and foregoes the recursive bias learning in RNN, while this recursive learning might be more important than commonly believed.

Dehghani et al. point out that re-introduce recurrence calculation in transformer maybe promote the performance of transformer, which is called as universal transformer \cite{22DBLP:conf/iclr/DehghaniGVUK19}, this model combines the advantage of transformer and RNN, organically combines the self-attention and recurrence learning mechanism. In recurrence process, Dehghani et al. make use of the ACT mechanism \cite{23graves2016adaptive} to decide when the recurrence process will halt. The experimental results of universal transformer demonstrate that the effectiveness of the combination of self-attention mechanism and recurrence mechanism.

Based on the achievements of universal transformer, we tend to work out a new framework of transformer Hawkes process based on the idea of universal transformer, we name it as universal transformer Hawkes process. We introduce the recurrent structure in transformer Hawkes process, and make our model to achieve Turing completeness compared with previous transformer model. Moreover, we add a convolution module to the position-wise-feed-forward-layer, which enhance the local perception ability of universal transformer Hawkes process. We conduct the experiments on multiple dataset to compare with state-of-the-art baselines, to validate the effectiveness of our model. We also demonstrate whether the additional RNN layers will have a positive impact on fitting mutual interdependence among the events in the sequence. In addition, to demonstrate the effectiveness of the ACT mechanism, we compare the performances of universal transformer with and without the ACT mechanism, and verify that the halting mechanism of dynamic iteration will make the model perform better overall. 

Our paper is organized as follows. In Section 2, we introduce the related work about Hawkes process and point process in view of the neural network. In Section 3, we are going to instruct our model in details, including structure, condition intensity function, prediction tasks and training process. At last, Section 4 lists our experimental results to illustrate the advantages of universal transformer Hawkes process and t ACT mechanism. At last, Section 5 concludes the article.

\section{Related work}

\subsection{Hawkes process}

Hawkes process has form shown as the following:

\begin{equation}
\label{eq1}
\lambda (t) = \mu (t) + \sum\limits_{i:t_i  < t} {\phi (t - t_i )} 	
\end{equation}

where $\mu (t)$   is background intensity function, indicates the background probability of event occurrence,  $\phi (t)$  is the impact function, which used to measure the historical event influence, and $\sum\limits_{i:t_i  < t} {\phi (t - t_i )} $  records the impact of all historical events on the current instant. Traditional Hawkes process model in Eq. 1 assumes the positive superposition of past historical impact. Until now, there are many variants of traditional Hawkes process.

Zhao et al. firstly make use of Hawkes process to model the Twitter data to predict the final number of reshares, which only have small relative error \cite{6zhao2015seismic}. Xu et al. use the non-parametric method to represent the Hawkes process, and utilize a corresponding learning algorithm to get the better performance of the model, and then the Granger causality on of IPTV user on watching program is obtained \cite{8xu2016learning}. Kobayashi and Lamhbiotte present a time-dependent Hawkes process, whose impact functions are time-dependent. They validate the method on Twitter data and prove that there is a systematic improvement to the previous method \cite{24kobayashi2016tideh}. Yang et al. develop an online learning method of Hawkes process based on the nonparametric method \cite{25yang2017online}. In 2018, Alan reviews and summarizes the application of the Hawkes process proposed by him in the financial field \cite{26hawkes2018hawkes}. Mohler presents a novel modulated Hawkes process, to quickly identify risks and trigger appropriate public safety responses in communities to prevent a range of social harm events, such as crime, drug abuse, traffic accident and medical emergencies \cite{27mohler2018improving}.

Zhang et al. \cite{4zhang2020survival} make significant improvement to the Hawkes process model, the Weibull background intensity is used instead of constant background intensity. Weibull background intensity is a kind of time-dependent background intensity, which can better describe the trend of the base possibility of the event over time. After verifying the effectiveness of this Weibull-Hawkes process and the corresponding learning algorithm, based on this new model, Zhang et al. analyze the failure sequence of the compressor station and list the trend of background probability of failures in the compressor station over time and the Granger causality between the failures, and some suggestions are made for the production of the compressor station.

\subsection{Transformer and Universal Transformer}

In 2017, Vaswani et al. propose transformer model \cite{18vaswani2017attention}, which makes full use of self-attention \cite{19DBLP:journals/corr/BahdanauCB14} module, and discards the CNN and RNN structure, achieves a great improvement in the field of sequence learning, such as natural language processing \cite{28chowdhary2020natural}.

However, recent research shows that the recurrent learning in RNN can have a greater role beyond imagination, thus, Dehghani et al. propose universal transformer \cite{22DBLP:conf/iclr/DehghaniGVUK19} which combine the recurrent learning and self-attention mechanism, moreover, in order to better allocate model computing resources, the ACT mechanism \cite{23graves2016adaptive} is introduced into models, then, this model achieves better results than transformers.

\subsection{Neural Hawkes Process}

In general, the neural network has a stronger nonlinear fitting ability than other models, especially RNN is better at learning sequence data than other neural networks, so it has been often used in the processing of sequence data such as speech and text.

Therefore, researchers also hope to use RNN to fit asynchronous event sequence data. Du et al. present RMTPP models \cite{11du2016recurrent}, to learn the history effect via RNN, including history event type and time-stamp. For the first time, this neural model abandons the overly strong assumptions of the Hawkes process and other point process models, and achieves greater improvements. Xiao et al. make use of two RNN to model the event sequence, one of them is used to model the background intensity and the other is used to model the impact of historical events \cite{13xiao2017modeling}. This method allows a black-box treatment for modeling the conditional intensity function, and end to end training can make model event sequences easier. Mei and Eisner propose a new form of LSTM \cite{12mei2017neural}, the continuous time LSTM, whose state can decay to another with time, and based on this LSTM they come up with the neural Hawkes process to model the asynchronous event sequence, which obtains better model performance.

While the shortcomings of RNN gradually appear, researchers tend to use better neural networks to model event sequences. Zhang et al. \cite{20zhang2020self} firstly utilize self-attention mechanism to get the Hawkes process of the sequence. And based on the achievement of transformer, and enlighten by \cite{18vaswani2017attention}, Zuo et al. utilize the encoder structure in transformer, and propose the transformer Hawkes process (THP) \cite{21DBLP:conf/icml/ZuoJLZZ20}, this model encodes and converts event sequence data into hidden representations, and then maps hidden representations into continuous conditional intensity functions.

\section{Proposed Model}

\begin{table}[!htbp]
	\caption{Nomenclature}
	\label{tb1}
	\resizebox{150mm}{93mm}{
	\begin{tabular}{cccc}
		\hline	
		Symbols	& Description & Variable names in ACT mechanism  & Size\\ \hline
		$S_e$	&The dataset of sequences & / & /  \\
		$s_n$	& The $n$-th sequence &/ & /  \\
		$I_n$	& The length of $n$-th sequence &/ & /  \\
		$N$	&The total number of sequences &/ & $\mathbb{N}^ +$  \\
		$C$	&The total number of type of events in sequences &/ & $\mathbb{N}^ +$  \\
		$t_i$	& The time stamp of $i$-th event &/ & $\mathbb{R}$  \\
		$c_i$	& The event type of $i$-th event &/ & $\mathbb{N}^ +$ \\
		$D$	&The model dimension of universal transformer & /& $\mathbb{N}^ +$  \\
		$\bm{K}$	& The embedding matrix of event type & /& $\mathbb{R}^ {D\times C}$  \\
		max\_n	&  The max iteration times of ACT mechanism & /& $\mathbb{N}^ +$  \\			
	$\mathbb{I}( \cdot )$	&Indicator function &/& /  \\						
		\textit{EncodingLayer}	& The encoding layer in ACT mechanism &/ & / \\ 
	$T_h$	& The threshold in ACT mechanism  & threshold & $\mathbb{R}$ \\
	$\bm{W}_p$	& The weight p in ACT mechanism & weight\_p & $\mathbb{R}^ {D\times 1}$ \\ 
	$S$	& The state variable in ACT mechanism  & state & $\mathbb{R}^ {I_n\times D}$ \\ 
	$P_S$	& The previous state variable in ACT mechanism  & previous\_state & 
	$\mathbb{R}^ {I_n \times D}$   \\ 
	$h_p$	& The halting probability variable in ACT mechanism &halting\_probability & $\mathbb{R}^ {D}$ \\ 
	$R_e$	&The remainders variable in ACT mechanism &remainders & $\mathbb{R}^ {D}$ \\ 
	$n$	& The upgrade times of state in  ACT mechanism  & n\_updatas & $\mathbb{R}^ {D}$ \\
	$S_r$	& The still running variable in ACT mechanism &still\_running & $\mathbb{R}^ {D}$ \\ 
	$n_h$	& The new halted variable in ACT mechanism &new\_halted & $\mathbb{R}^ {D}$ \\
	$W$	& The update weight of ACT mechanism &update\_weight & $\mathbb{R}^ {D}$ \\ 
	$L$	&he number of multi-head attention & /& $\mathbb{R}$ \\ 
	$\left\{ {{\bm{A}}_l } \right\}_{l = 1}^L $	& The number of multi-head attention &/ & $\mathbb{R}^ {D_V}$ \\ 
	${D_K}$	&The dimension of query and key vector &/ & $\mathbb{N}^ +$ \\ 
	${D_V}$	& The dimension of value vector &/ & $\mathbb{N}^ +$ \\ 
	$\left\{ {{\bm{Q}}_l } \right\}_{l = 1}^L$	&The query variable of multi-head attention &/ & $\mathbb{R}^{I_n  \times D_K }$ \\ 
	$\left\{ {{\bm{K}}_l } \right\}_{l = 1}^L$	&The key variable of multi-head attention &/ & $\mathbb{R}^{I_n  \times D_K }$ \\ 
	$\left\{ {{\bm{V}}_l } \right\}_{l = 1}^L$	&The value variable of multi-head attention &/ & $\mathbb{R}^{I_n  \times D_V }$ \\ 
	$\left\{ {{\bm{W}}_Q^l } \right\}_{l = 1}^L$ 	&The query matrix of multi-head attention &/ &$\mathbb{R}^{D  \times D_K }$  \\ 
	$\left\{ {{\bm{W}}_K^l } \right\}_{l = 1}^L$ 	&The key matrix of multi-head attention &/ & $\mathbb{R}^{D \times D_K }$ \\
	$\left\{ {{\bm{W}}_v^l } \right\}_{l = 1}^L$ 	&The value matrix of multi-head attention &/ &$\mathbb{R}^{D  \times D_K }$  \\  
	$W_{multi}$	& The aggregation matrix of multi-head attention &/ &  $\mathbb{R}^{LD_V  \times D }$\\ 
 $\bm{A}'$&Obtained from $\bm{A}$  passes through fully connected layer FC1 &/ & $\mathbb{R}^{I_n  \times D }$ \\ 
  $\bm{A}''$& Obtained from $\bm{A}'$ passes through CNN layer and fully connected layer FC2 &/ & $\mathbb{R}^{I_n  \times D_H }$ \\ 
 $D_{RNN}$& The dimension of RNN in the model &/ &  $\mathbb{N}^ {+}$ \\ 
$S'$ & Obtained from , the output of ACT mechanism, passes through fully connected layer FC3 &/ & $\mathbb{R}^{I_n  \times D_{RNN}}$ \\ 
$S''$ & Obtained from  passes through RNN and fully connected layer FC4 &/ & $\mathbb{R}^{I_n  \times D}$ \\ 
$\bm{H}$ &The hidden representation of sequence &/ & $\mathbb{R}^{I_n  \times D}$ \\ 
$\bm{h}(t_i)$ & The hidden representation of \textit{i}-th event  &/ & $\mathbb{R}^{ D}$ \\
$\mathcal{H}_t$	 & The previous history at time \textit{t} & / & / \\
$b_c$		& The background intensity of event type \textit{c} &/ & $\mathbb{R}$ \\
${\alpha _c } $	& Thecontinuous parameter of conditional intensity function of event type &/ &  $\mathbb{R}^{1\times D}$ \\
${\bm{w}}_c^T $& Historical weight parameter of conditional intensity function of event  type &/ & $\mathbb{R}^{1\times D}$ \\
$\bm{W}_{time}$		& The prediction parameter of time-stamp&/ & $\mathbb{R}^{1\times D}$ \\
$\bm{W}_{type}$		&The prediction parameter of event type &/ & $\mathbb{R}^{C\times D}$ \\ \hline
				
	\end{tabular}}
\end{table}

For sequences of asynchronous events, we need to determine what model they have. The symbols used in the paper are shown in Table 1, and in general, we formulate it as following: assume that there are $N$ sequences in the dataset of the asynchronous events, represented as $S_e  = \{ s_n \} _{n = 1}^N$ , and for each sequence, note that their lengths are not the same, for the  $n$-th sequence $s_n  = \{ t_i ,c_i \} _{i = 1}^{I_n } $ , its length is $I_n$ , each sequence $s_n$  is composed with $I_n$ tuples, $t_i$ is the time-stamp of  $i$-th event, and $c_i\in C$ is the corresponding event type. In the forward phase, for arbitrary sequence $s_n  = \{ t_i ,c_i \} _{i = 1}^{I_n } $ , the entire sequence could be directly input to the universal transformer, supposing the model dimension is $D$ , then all these events in the sequences (including their time-stamp) can be represent by their corresponding  $D$-dimensional vectors, then this sequence can be described by the hidden representation of event sequence ${\bm{H}} \in \mathbb{R}^{D \times I_N } $ . The continuous conditional intensity function can be calculated by ${\bm{H}} \in \mathbb{R}^{D \times I_N } $ and the equation we proposed in section 3.2.

\subsection{Universal Transformer Hawkes Process}

First, we need to map the asynchronous event sequence into the temporal encodings, which denote the occurring times of events, and event-type encodings. According to point process theory, the time-stamps of events in the event sequences are the coordinate on the temporal axis, in other words, time-stamps are temporal position of events. Analogous to the text sequence, the time-stamps in the event sequence are equivalent to the position of the words, and the event type embedding vectors are equivalent to the semantic vectors of the words. Thus, for the temporal encodings, we can leverage the position encoding approaches in nature language processing to encode the time-stamps of events, similar as [18,21], the position encoder for the occurring time-stamp of event is shown as Eq.2:

\begin{equation}
\label{eq2}
[{\bm{x}}(t_i )]_j  = \left\{ {\begin{array}{*{20}c}
	{\cos \left( {t_i /10000^{\frac{{j - 1}}{D}} } \right),{\rm{if}}\:j\:{\rm{is}}\:{\rm{odd}},}  \\
	{\sin \left( {t_i /10000^{\frac{j}{D}} } \right),{\rm{if}}\:j\:{\rm{is}}\:{\rm{even}}.}  \\
	\end{array}} \right.	
\end{equation}

Thus, for the time-stamp $t_i$ of event, its corresponding temporal encoding is $
{\bm{x}}(t_i ) \in \mathbb{R}^D $ , where $D$ is the model dimension of universal transformer, and in order to get event-type encoding, we set embedding matrix ${\bm{K}} \in \mathbb{R}^{D \times C} $, and each kind of event is corresponding to an one-hot encoding ${\bm{c}}_i  \in \mathbb{R}^C $ , then we can get the event-type encoding $
{\bm{Kc}}_i  \in \mathbb{R}^D $ . We define that $\bm{X}^T$  and $({\bm{KC}}_n )^T$ are the corresponding temporal encoding and event-type encoding of the sequence, where $
{bm{X}} = \{ {bm{x}}(t_1 ),{bm{x}}(t_2 ),...,{bm{x}}(t_{I_n } )\}  \in \mathbb{R}^{D \times I_n } $ and ${bm{C}}_n  = [c_1 ,c_2 ,...,c_{I_n } ] \in \mathbb{R}^{C \times I_n } $ . 

In order to get the hidden representation of event sequence, we need to input the above temporal and event-type encodings into the recurrent part in universal transformer, which is shown as Fig. 1 and Fig. 2.

\begin{figure}[!htbp]
	\centering
	\includegraphics[scale=0.7]{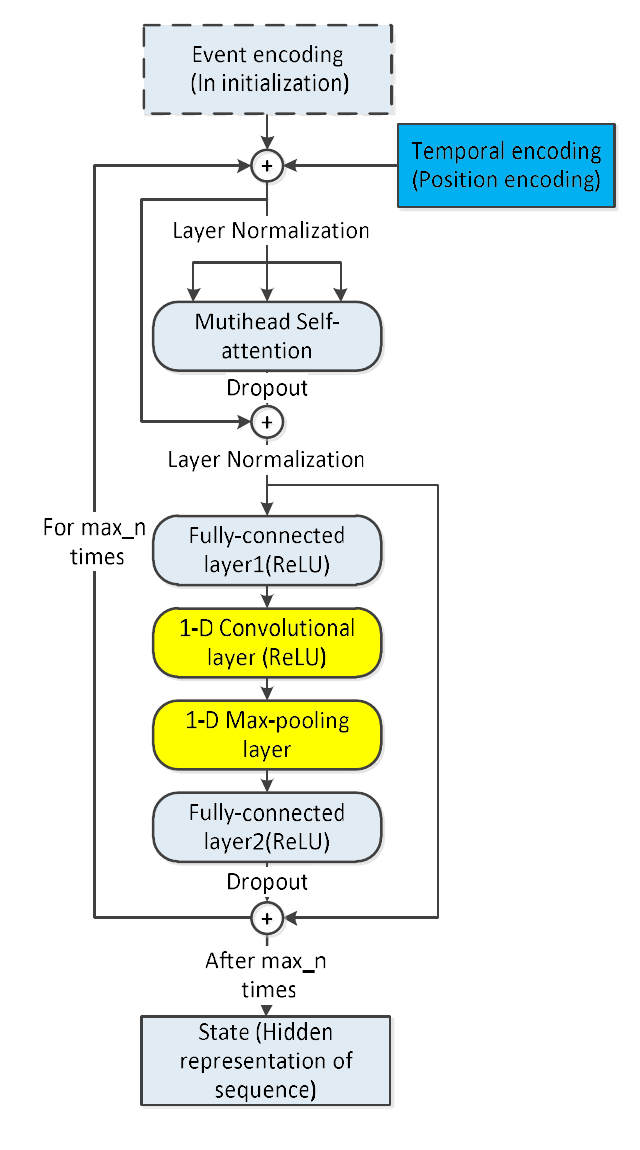}
	\caption{Schematic diagram of universal transformer with pure recurrence, the model simply uses the hidden state as a recurrent variable }
	\label{fig1}
\end{figure}

\begin{figure}[!htbp]
	\centering
	\includegraphics[scale=0.7]{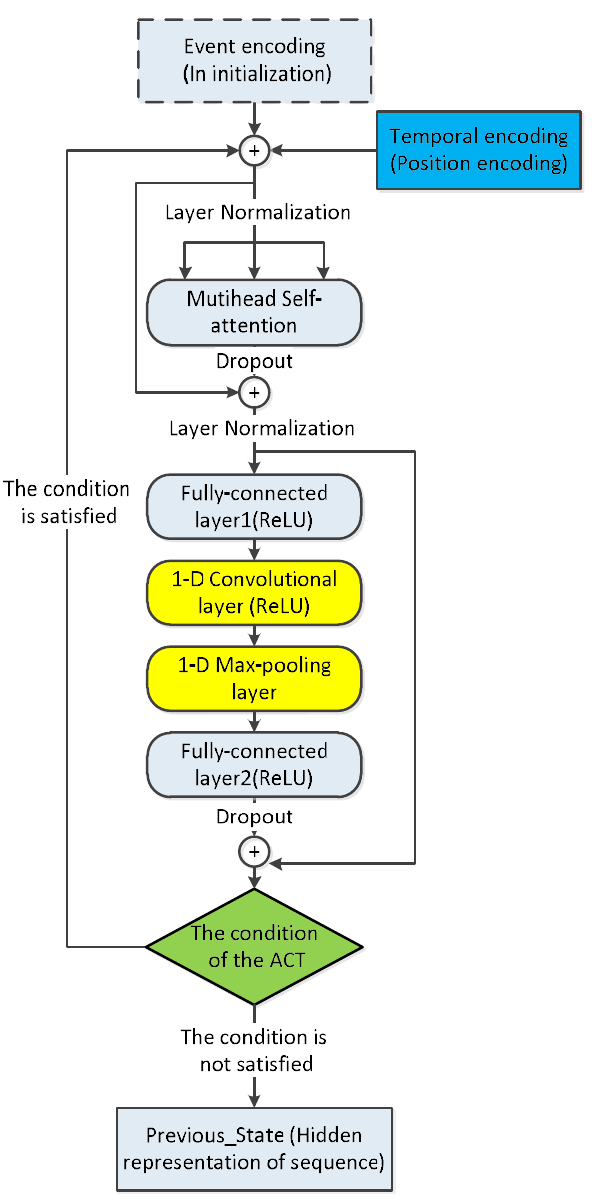}
	\caption{Schematic diagram of universal transformer based on ACT algorithm, in the model, ACT algorithm provides a measure of how well of learning of different elements in the variable.}
	\label{fig1}
\end{figure}

Different with traditional transformer, for universal transformer, the temporal and event-type encodings will be alternatively and recursively refined by attention and convolutional layers until a certain stop condition is met, then the final hidden representation will be acquired. In universal transformer with pure recurrence illustrated in Fig. 1, the hidden state will iterate the predetermined number of times, and the final hidden state is set as the hidden representation of event sequence. The corresponding implementation process is shown as the Algorithm 1:

\begin{algorithm}
	\renewcommand{\algorithmicrequire}{\textbf{Input:}}
	\renewcommand{\algorithmicensure}{\textbf{Output:}}
	\caption{Universal Transformer with pure recurrence.}
	\label{alg1}
	\begin{algorithmic}[2]
		\REQUIRE The maximum number of iterations: max\_n, event-type encoding $({\bm{KC}}_n )^T $, temporal encoding (position encoding) $\bm{X}^T$  and \textit{EncodingLayer} in model.
		\ENSURE Hidden representation of event sequence  ${\bm{H}} \in \mathbb{R}^{D \times I_N } $ (State) 
		\STATE  Initialize state $n\leftarrow0$, $ {\bm{S}} \leftarrow ({\bm{KC}}_n )^T $ .
		\STATE \textbf{while} $(n<\rm{max\_n})$ \textbf{do}
		\STATE ${\bm{S}} \leftarrow {\bm{S}} + {{ }}{\bm{X}}^T $
		\STATE ${\bm{S}} \leftarrow output\;of\;{\rm{the}}\;i{\rm{ - th }}\;{\rm{encoding\;layer}}({\bm{S}})$
		\STATE \textbf{end while}	
		\STATE \textbf{return} ${\bm{H}} \leftarrow {\bm{S}}$
	\end{algorithmic}
\end{algorithm}

In iterative process, some symbols, such as words and phonemes, in our application scenario they are the temporal and event-type encodings, are usually more ambiguous than other elements in the sequence, the more ambiguous these symbols are, and the more refining processes are required. However, the above algorithm simply refines equally among each element in the sequence, it is easy to see that this mechanism is too rough and doesn’t meet the requirement we mentioned.

Thus, similar as \cite{22DBLP:conf/iclr/DehghaniGVUK19}, we adopt the ACT mechanism \cite{23graves2016adaptive} in our proposed universal transformer Hawkes process, which is a mechanism can dynamically adjust the number of refinedness for each symbol in the sequence. The ACT mechanism updates the symbols in the sequence according to their state parameter. Since ACT method is used in speech recognition and other scenarios and cannot be directly applied to event sequence data, we are going to make some certain modifications to the concrete implementation procedure of the ACT mechanism. The modified ACT mechanism is illustrated in Algorithm 2, and the information flow diagram of ACT mechanism in Universal transformer is shown in Fig. 3.

\begin{algorithm}
	\renewcommand{\algorithmicrequire}{\textbf{Input:}}
	\renewcommand{\algorithmicensure}{\textbf{Output:}}
	\caption{Universal Transformer based on ACT mechanism.}
	\label{alg2}
	\begin{algorithmic}[2]
		\REQUIRE The maximum number of iterations: max\_n, event-type encoding $({\bm{KC}}_n )^T $, temporal encoding (position encoding) $\bm{X}^T$, threshold $T_h$ and \textit{EncodingLayer} in model.\\ \textbf{Paremeter}:${\bm{W}}_p  \in \mathbb{R}^{D \times 1} $			
		\ENSURE Hidden representation of event sequence  ${\bm{H}} \in \mathbb{R}^{D \times I_N } $ (State) 
		\STATE  Initialize state $n \leftarrow {0},R_e  \leftarrow 0,h_p \leftarrow 0,P_s \leftarrow 0	$ and $ {\bm{S}} \leftarrow ({\bm{KC}}_n )^T $ .
		\STATE \textbf{while} there is any element in  $\left( {(h_p  > T_h )\& (n < \max \_n)} \right)$ is true, \textbf{do}
		\STATE ${\bm{S}} \leftarrow {\bm{S}} + {{ }}{\bm{X}}^T $
		\STATE $p \leftarrow Sigmoid (S{\bm{W}}_p )$
		\STATE $S_r  \leftarrow \mathbb{I}(h_p  < 1.0)$
		\STATE $N_h  \leftarrow (h_p  + (p*\mathbb{I}(S_r  > T_h ))*S_r$
		\STATE$S_r  \leftarrow (h_p  + (p*\mathbb{I}(S_r  \geqslant T_h ))*S_r $
		\STATE$	h_p  \leftarrow h_p  + p*S_r $		
		\STATE$	R_e  \leftarrow R_e  + N_h *(1 - h_p )	$	
		\STATE$	h_p  \leftarrow h_p  + N_h *R_e $	
		\STATE$	n \leftarrow n + S_r  + N_h $
		\STATE$	W \leftarrow p*S_r  + N_h *R_e 	$
		\STATE$	S \leftarrow EncodingLayer(S)	$	
		\STATE$	P_S  \leftarrow S*W + P_S *(1 - W)	$
		\STATE \textbf{end while}	
		\STATE \textbf{return} ${\bm{H}} \leftarrow {{P_S}}$
	\end{algorithmic}
\end{algorithm}

\begin{figure}[!htbp]
	\centering
	\includegraphics[scale=0.7]{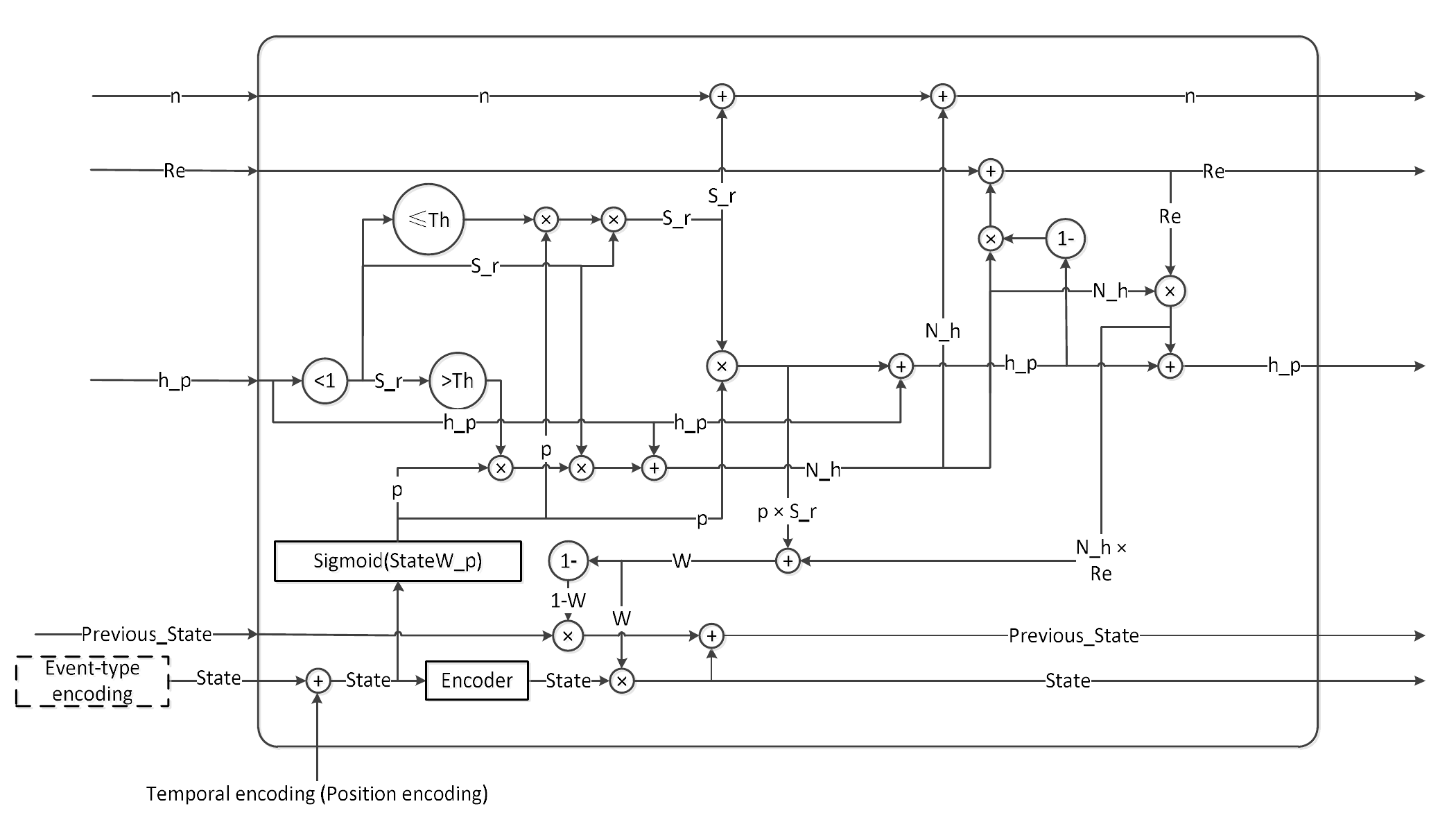}
	\caption{Diagram of universal transformer based on ACT mechanism, the figure shows the information flow and update process of different variables }
	\label{fig3}
\end{figure}

We can see that universal transformer without ACT mechanism and RNN have obvious similarities, in both models, there is only state calculated recursively in the iteration, and universal transformer with ACT mechanism is more similar to LSTM, because there are multiple variables to be updated in the recurrence, and the updated weights are to some extent consistent with the forgetting gate in LSTM. These facts show that universal transformer with ACT mechanism will have a stronger learning ability than universal transformer without ACT mechanism.

\begin{figure}[!htbp]
	\centering
	\includegraphics[scale=0.7]{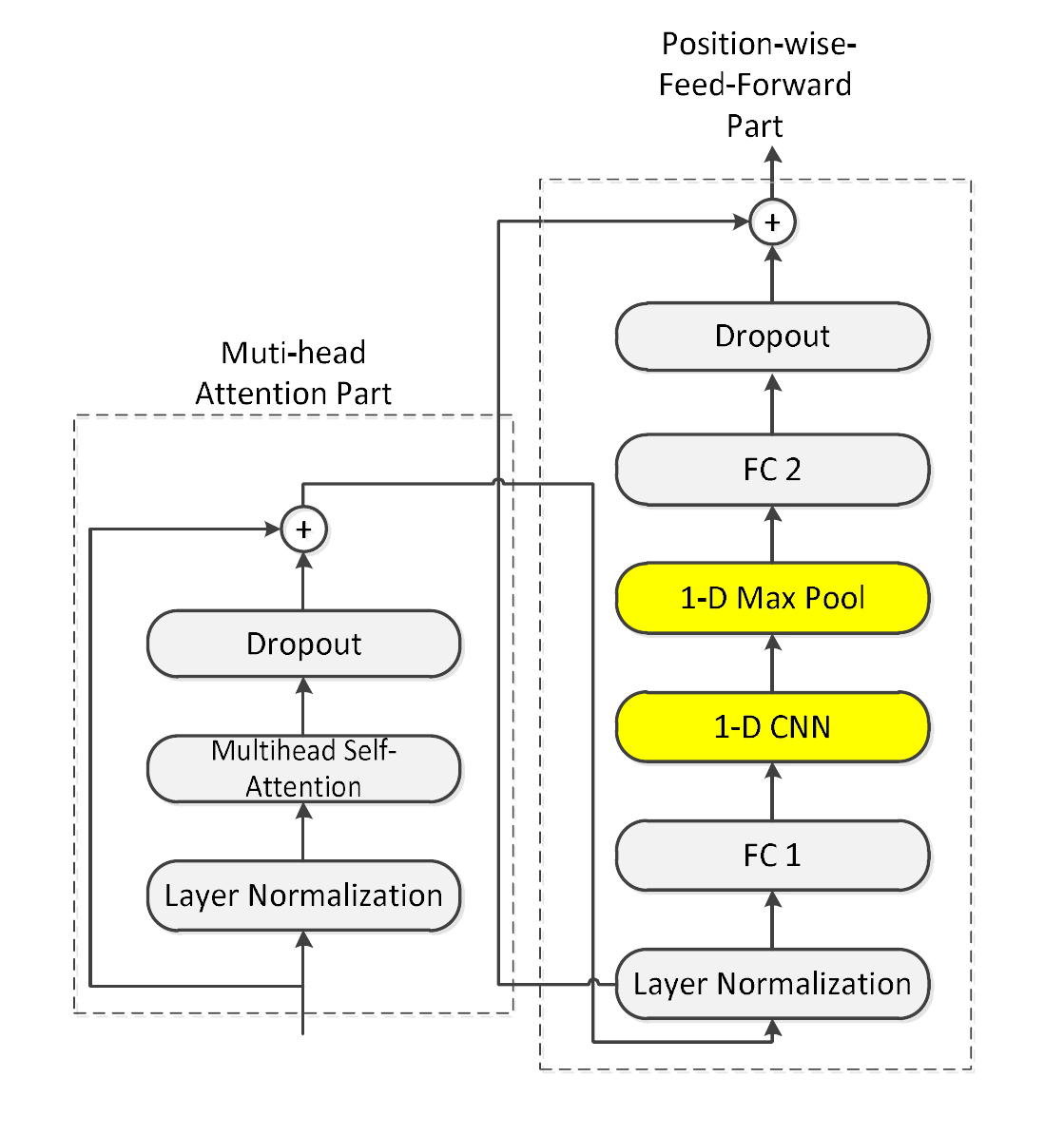}
	\caption{The diagram of encoding layer in universal transformer, the self-attention mechanism of universal transformer is similar to the normal transformer, and we modify position-wise-feed-forward part, add a convolutional layer between the two fully connected layers to enhance the local perception ability of the model. }
	\label{fig4}
\end{figure}

In both algorithms, we all are going to feed the state  in Algorithm 1 and 2 into the encoding layer, whose flow-chart is illustrated in Fig. 4, encoding layer is composed with two parts, i.e., multi-head attention part and position-wise-feed-forward part. In encoding layer, we also utilize the commonly used techniques in transformer to relieve the difficulty in training, such as the residual connection \cite{29he2016deep} and layer normalization \cite{30ba2016layer}.

In multi-head attention part, we also adopt dot-product attention, which can be written as Eq.3:

\begin{equation}
{\bm{A}} = Softmax\left( {\frac{{{\bm{QK}}^T }}
	{{\sqrt {D_K } }}} \right){\bm{V}}
\end{equation}

where

\begin{equation}
{\bm{Q}} = {\bm{SW}}_Q ,{\bm{K}} = {\bm{SW}}_K ,{\bm{V}} = {\bm{SW}}_V 
\end{equation}

state $S$ is the input to the encoding layer, $\bm{Q}$,$\bm{K}$,and $\bm{V}$  are respectively query, key and value matrices, which are obtained by different linear transformations ${\bm{W}}_Q ,{\bm{W}}_K  \in \mathbb{R}^{D \times D_K } $
, $ {\bm{W}}_V  \in \mathbb{R}^{D \times D_V } $ of state $S$ . And in order to improve the diversity and expressive ability of model, we make use of multi-head self-attention to make the model perform better. If we utilize L heads self-attention, then, we can get $L$  outputs ${\bm{A}}_1 ,{\bm{A}}_2 ,...,{\bm{A}}_L 
$ , which are computed by different sets of matrices $\left\{ {{\bm{W}}_Q^l ,{\bm{W}}_K^l ,{\bm{W}}_V^l } \right\}_{l = 1}^L $  . The formula for multi-head attention is shown as Eq.5:

\begin{equation}
{\bm{A}} = \left[ {{\bm{A}}_1 ,{\bm{A}}_2 ,...,{\bm{A}}_L } \right]{\bm{W}}_{multi} 
\end{equation}

In Eq.5, ${\bm{W}}_{multi}  \in \mathbb{R}^{LD_V  \times D}$  is the aggregation matrix. It is worth noting that the events in the sequence occur in sequential order, thus, it has the characteristic called as “future invisibility”, we need to block the impact of future events on current time. We utilize the masked self-attention mechanism similar as \cite{18vaswani2017attention}. For example, we set ${\bm{E}} = {\bm{QK}}^T $ , the masked self-attention mechanism will set the element in the  $i$-th row in $\bm{E}$ ranged in ${\bm{E}}(i,i + 1),{\bm{E}}(i,i + 2),..,{\bm{E}}(i,D)$
to negative infinite, which will ensure that the self-attention weight of future events obtained through softmax function is 0.

After getting the attention output ${\bm{A}}$ it will be fed into position-wise-feed-forward part, which is shown in the right part of Fig. 4. To improve model’s local perception of the sequence, we introduce the CNN module in position-wise-feed-forward part, more specifically, fully connected layer FC1 transforms attention output ${\bm{A}}$  into a higher-dimensional linear matrix ${\bm{A'}} \in \mathbb{R}^{I_n  \times D_H } $ . Then it pass through the convolutional layer, nonlinear layer, and pooling layer, and fully connected layer FC2 will convert it to the original dimension ${\bm{A''}} \in \mathbb{R}^{I_n  \times D}$.

After the Algorithm 1 or 2 completely stop, we will get the output $H$, and at last,  will be as the input of postprocessing part, which is constituted with fully connected layer FC3, RNN layer and final fully connected layer FC4. This structure is motivated by \cite{21DBLP:conf/icml/ZuoJLZZ20},\cite{31wang2019language}, and in fact the additional RNN can make model better fit the sequential data, in a nutshell, the FC3 transforms input  $S$ or $P_S$  to $S' \in \mathbb{R}^{I_n  \times D_{{\rm{RNN}}} } $  and $S'$  will be recurrently iterated by the RNN, note that the types of  RNN can be selected as LSTM or GRU, then the FC4 converts $S'$  into $S'' \in \mathbb{R}^{I_n  \times D } $ , at last, $S''$  will be assigned to the hidden representation ${\bm{H}} \in \mathbb{R}^{I_n  \times D} $ of the sequence of events. The effectiveness of postprocessing part in UTHP is decided by the characteristic of the dataset, and we will discuss this issue in subsection 4.3.2 and analyzed it in experiments.

\subsection{Conditional Intensity Function}

In general, point processes are supposed to be described by their corresponding conditional intensity functions. However, after encoding operation of universal transformer, we can only get the discrete conditional intensity established by hidden representation of event sequence. Therefore, we need to construct the continuous conditional intensity function on the foundation of hidden representation of event sequence.

For every kind of event, we denote $\lambda _c (t\left| {\mathcal{H}_t } \right.)$ as corresponding conditional intensity function of UTHP, and here $\mathcal{H}_t  = (t_i ,c_i ):t_i  < t$ is the history of past events at time \textit{t}. Similar as \cite{21DBLP:conf/icml/ZuoJLZZ20}, the form of conditional intensity function is shown as Eq.6:

\begin{equation}
\lambda _c (t\left| {\mathcal{H}_t } \right.) = f(b_c  + \alpha _c \frac{{t - t_i }}
{{t_i }} + {\bm{w}}_c^T {\bm{h}}(t_i ))	
\end{equation}

In Eq.6, $t \in [t_i ,t_{i + 1} )$, and $b_c$  is the background intensity, indicates the possibility of an event occurring without considering historical information. $\alpha _c \frac{{t - t_i }}{{t_i }}$  is continuous conditional intensity function in the interval $[t_i ,t_{i + 1} )$ , where $\alpha _c$ is predefined parameter and can be modified to trainable parameter in subsequent improvements, and ${\bm{w}}_c^T {\bm{h}}(t_i ))$  represents the history impact to the conditional intensity function. 
$f( \cdot )$is the softplus function, which is a smooth improvement of the ReLU nonlinear function. Softness parameter of softplus function is $\beta$ , using this nonlinear function makes the conditional intensity function smoother and make the new model have a better expressive ability. And the overall conditional intensity function for the whole sequence is given as follow:

\begin{equation}
\lambda (t\left| {\mathcal{H}_t } \right.){\rm{  =  }}\sum\limits_{c = 1}^C {\lambda _c (t\left| {\mathcal{H}_t } \right.)} 
\end{equation}

\subsection{Prediction}

In point process theory, prediction is one of the most important tasks. For instance, it can be used to predict when does the patient develop which disease, and when and what events will happen in social networks. After obtaining the conditional intensity function, in view of \cite{32daley2007introduction}, we can predict the possible types and occurrence times of future events according to conditional intensity function and Eq.8:

\begin{equation}
\begin{array}{l}
p(\left. t \right|{\cal H}_t ) = \lambda (\left. t \right|{\cal H}_t )\exp t( - \int_{t_i }^t {\lambda (\left. s \right|{\cal H}_t )} ds) \\ 
\hat t_{i + 1}  = \int_{t_i }^\infty  {t\cdot p(\left. t \right|{\cal H}_t )dt}  \\ 
\hat c_{i + 1}  = \mathop {\arg \max }\limits_c \frac{{\lambda _c (\hat t_{i + 1} |{\cal H}_{i + 1} )}}{{\lambda (\hat t_{i + 1} |{\cal H}_{i + 1} )}} \\ 
\end{array}
\end{equation}

The existence of the integral of the conditional intensity function in Eq.8 makes the prediction become difficult, although we can use the Monte Carlo sampling method \cite{33robert2013monte} to calculate it, this is still very inefficient. Due to the powerful fitting ability of neural network, we make use of neural work to predict the future event type and time-stamps, which can be shown as Eq.9:

\begin{equation}
\begin{array}{l}
\hat t_{i + 1}  = {\bm{W}}_{time} {\bm{h}}(t_i ) \\ 
\widehat{\bf{p}}_{i + 1}  = Softmax({\bm{W}}_{type} {\bm{h}}(t_i )) \\ 
\hat c_{i + 1}  = \mathop {\arg \max }\limits_c {\bm{\hat p}}_{i + 1} (c) \\ 
\end{array}
\end{equation}

According to \cite{32daley2007introduction}, given the sequence $s_n  = \{ t_i ,c_i \} _{i = 1}^{I_n } $ , and conditional intensity function, we can get its log-likelihood expression:

\begin{equation}
L(s_n ) = \sum\limits_{i = 1}^{I_n } {\log } \,\lambda (\left. {t_i } \right|\mathcal{H}_i ) - \int_{t_1 }^{t_{I_n } } {\lambda (\left. t \right|\mathcal{H}_t )} dt
\end{equation}

Assuming there are $N$  sequences, then the model parameters can be solved by maximum log-likelihood principle:

\[
\max \sum\nolimits_{n = 1}^N {L(s_n )} 
\]

However, there is difficulty in solving  $\Lambda  = \int_{t_1 }^{t_{I_n } } {\lambda \left( {\left. t \right|\mathcal{H}_t } \right)} dt$ because we use universal transformer to obtain the conditional intensity function, it is hard to obtain the closed-form of $\Lambda$ , fortunately, we have two alternative methods to solve it. The first method is the Monte Carlo sampling integration method \cite{33robert2013monte}, which is shown in Eq.11:

\begin{equation}
\label{eq11}
\hat \Lambda _{MC}  = \sum\limits_{i = 2}^L {\left( {t_i  - t_{i - 1} } \right)} (\frac{1}
{M}\sum\limits_{m = 1}^M {\lambda (u_m )} )
\end{equation}

where $u_m \sim U(t_{i - 1} ,t_i )$ , and  $u_m$ is sampled from the uniform distribution in support set $[Ut_{i - 1} ,t_i ]$. $\hat \Lambda _{MC}$  calculated by this method is an unbiased estimate of $\Lambda$. The second method is numerical analysis method \cite{34stoer2013introduction}, for instance, based on the trapezoidal rule, we can get the estimate as shown in Eq.12:

\begin{equation}
\hat \Lambda _{NI}  = \sum\limits_{i = 2}^{I_n } {\frac{{t_i  - t_{i - 1} }}
	{2}} (\lambda (\left. {t_i } \right|\mathcal{H}_i ) + \lambda (\left. {t_{i - 1} } \right|\mathcal{H}_{i - 1} ))
\end{equation}

In spite of the bigger estimate error than the Monte Carlo method, $\hat \Lambda _{NI}$  is still a valid approximation of  $\Lambda$ . After dealing with the solution of  $\Lambda$ , we are going to define the prediction loss of time-stamp and event type, for a sequence of length $I_n$ , we will make $I_n-1$  predictions of time-stamp and event type, in other words, we will not predict the first event. Therefore, the predict loss of next time-stamp and event prediction for sequence $s_n$ is shown as Eq.13 and 14:

\begin{equation}
L_{time} (s_n ) = \sum\nolimits_{i = 2}^{I_n } {(t_i  - \hat t_i )^2 } 
\end{equation}

\begin{equation}
L_{type} (s_n ) = \sum\nolimits_{i = 2}^{I_n } { - {\mathbf{c}}} _i^T \log (\widehat{\mathbf{p}}_i )
\end{equation}

where $\bm{c}_i$  is one-hot encoding with the ground-truth event type . In summary, given the dataset including the sequences$\{ s_n \} _{n = 1}^N$ , we need to solve:

\begin{equation}
\min \sum\limits_{n = 1}^N { - L(s_n )}  + \alpha _{type} L_{type} (s_n ) + \alpha _{time} L_{time} (s_n )
\end{equation}

where $\alpha _{type}$ and $\alpha _{time}$ are hyper-parameters, which will help keep training stable. This objective function can be efficiently solved by stochastic gradient optimization algorithm, such as adaptive moment estimation (ADAM) \cite{35kingma2014adam}, and we use the default parameters of ADAM: learning rate=0.0001, betas are 0.9 and 0.999, eps = $10^{-8}$, weight\_decay = 0, for all these experiments, when we solves log-likelihood of models, we set $\alpha _{type}=0$ and $\alpha _{time}=0$ , and when performing prediction tasks, we fix $\alpha _{type}=1$ and $\alpha _{time}=0.01$  .

\section{Experiments}

We compare our model with three baselines on six events sequence datasets, we evaluate these models by per-event-loglikelihood (in nats), root mean square error (RMSE) and event prediction accuracy on held-out test sets, we first introduce the details of the data set and baselines, and then list our experimental results.

\subsection{Datasets}
In this subsection, we utilize benchmark six datasets of event sequence to conduct experiments, Table 2 introduces the characteristics of each dataset.

% \usepackage{multirow}
\begin{table}[]
	\centering
	\caption{Characteristics of datasets. Each row corresponds to the description of this data set, including the total type number of events, the smallest, average, and largest sequence length.}
	\begin{tabular}{ccccc}
		\hline
		\multirow{2}{*}{Dataset} & \multirow{2}{*}{C} & \multicolumn{3}{c}{Sequence Length} \\ \cline{3-5} 
		&                    & Min       & Aver.       & Max       \\ \hline
		Synthetic                & 5                  & 20        & 60          & 100       \\
		Retweets                 & 3                  & 50        & 109         & 264       \\
		MemeTrack                & 5000               & 1         & 3           & 31        \\
		MIMIC-II                 & 75                 & 2         & 4           & 33        \\
		StackOverflow            & 22                 & 41        & 72          & 736       \\
		Financial                & 2                  & 829       & 2074        & 3319      \\ \hline
	\end{tabular}
\end{table}

\textbf{Synthetic} \cite{12mei2017neural}: Mei generates a set of synthetic sequences based on Hawkes process with random sampling the parameters and thinning algorithm.

\textbf{Retweets} \cite{6zhao2015seismic}: This dataset contains lots of sequences of tweets, every sequence corresponds to an origin tweet (for instance, the original content some user post) and its following retweets. The time and label of user of each retweet is recorded in the sequence, and users are divided into three categories based on the number of their followers: “small”, “medium”, and “large”.

\textbf{MemeTrack} \cite{36leskovec2014snap}: This dataset contains 42000 different meme posting in 5000 websites in the duration of 10 months, each sequence represents this meme from its birth to its final disappearance, and each event in the sequence corresponds to a time stamp and a website id.

\textbf{StackOverflow} \cite{36leskovec2014snap}: StackOverflow is a famous programming question-answering website. StackOverflow rewards users with badges to encourage them to participate in community activities, meanwhile, the same badge can be given to the same user, this dataset includes the lots of users reward histories during two years, which are treated as sequences, and each event in sequences indicates the acquisition of badge.

\textbf{Electrical Medical Records} \cite{37johnson2016mimic}: MIMIC-II dataset contains patients’ visit record to a hospital’s ICU in the course of seven years. Each patient’s record is treated as a sequence, and each event contains the corresponding time-stamp and diagnosis.

\textbf{Financial Transactions} \cite{11du2016recurrent}: This financial dataset contains large number of short-term transaction records of a stock in one day. The operation of each transaction is recorded as the event, and this dataset contains several long sequences, which only have two kinds of events: “buy” and “sell”, and the unit of the time-stamp is millisecond.

\subsection{Baselines}

\textbf{RMTPP} \cite{11du2016recurrent}: Du et al. come up with a recurrent network, which can model the event types and time-stamp in the sequence by embedding history to the vector.

\textbf{NHP} \cite{12mei2017neural}: Mei and Eisner present the neural Hawkes process based on the continuous-time LSTM, which has decay property of history events impact.

\textbf{THP} \cite{21DBLP:conf/icml/ZuoJLZZ20}: On the foundation of existing achievement of transformer, Zuo et al. propose the transformer Hawkes process, which achieves state-of-the-art performance. We get the following experimental results based on the model and the hyper-parameter they provide.

For NHP model, we also use ADAM optimizer with same hyper-parameter i.e., we set learning rate=0.0001, betas are 0.9 and 0.999, eps =$10^{-8}$ , weight\_decay = 0, and the number of hidden nodes of the continuous-time LSTM is 64, for Synthetic, Retweets and MemeTrack dataset, we set batch size to 16. The hyper-parameter configurations of THP and UTHP are given in section 4.5.

\subsection{Experimental results and comparison}

In this part, we are going to compare the performance of UTHP and baselines on datasets of multiple event sequences. First, we utilize loglike (per-event-loglikelihood) as the measurement of models’ performance on Synthetic, Retweets and MemeTrack datasets, which is shown in Table 3:

\begin{table}[]
	\centering
	\caption{The loglike on three dataset, from left to right, each column is a different model, and each row is a different dataset}
	\begin{tabular}{ccccc}
		\hline
		Datasets  & RMTPP            & NHP   & THP   & UTHP  \\ \hline
		Synthetic & \textbackslash{} & -1.33 & \textbf{0.834} & 0.796 \\
		Retweets  & -5.99            & -5.06 & \textbf{-4.69} & -4.75 \\
		MemeTrack & -6.04            & -6.23 &\textbf{ 0.68}  & 0.31  \\ \hline
	\end{tabular}
\end{table}

From Table 3, we can see that transformer models achieve obvious improvement than previous one, and although our UTHP model does not reach the state-of-the-art level, it still achieves second best. After this, we compare the performance of different models on complex datasets, including StackOverflow, MIMIC-II and Financial datasets. Because of the importance of event and time prediction of point process, while we compare the performance of models on these datasets, the multiple metric criteria will be adopted, involving per-event-log-likelihood, the predict accuracy for event and RMSE of time-stamp prediction. For transformer based model, we utilize Eq. 9 to make predictions, and for RNN based model, Eq. 10 is used. Table 4, Table 5 and Table 6 summarize the experimental results on these datasets.

\begin{table}[]
	\centering
	\caption{Different models’ performance on StackOverflow dataset, each row is the different metric criteria, including accuracy, RMSE and loglike (per-event-loglikelihood)}
	\begin{tabular}{ccccc}
		\hline
		StackOverflow & RMTPP & NHP   & THP   & UTHP  \\ \hline
		Accuracy      & 45.9  & 46.3  & 46.8  & \textbf{46.9}  \\
		RMSE(d)       & 9.78  & 9.83  & 4.99  & \textbf{4.42 } \\
		Loglike       & -2.60 & -2.55 & -0.56 & \textbf{-0.55 }\\ \hline
	\end{tabular}
\end{table}

\begin{table}[]	
	\centering
	\caption{Different models’ performance on MIMIC-II dataset.}
	\begin{tabular}{ccccc}
		\hline
		MIMIC-II & RMTPP & NHP   & THP   & UTHP  \\ \hline
		Accuracy & 81.2  & 83.2  & 83.2  & \textbf{84.4}  \\
		RMSE(d)  & 6.12  & 6.13  & 0.86  & \textbf{0.85}  \\
		Loglike  & -1.35 & -1.38 & \textbf{-0.14} & -0.18 \\ \hline
	\end{tabular}
\end{table}

\begin{table}[]
	\centering
	\caption{Different models’ performance on Financial dataset.}
	\begin{tabular}{ccccc}
		\hline
		Financial & RMTPP & NHP   & THP     & UTHP    \\ \hline
		Accuracy  & 61.95 & 62.20 & 62.23   & \textbf{62.52}   \\
		RMSE(s)   & 1.56  & 1.56  & 0.02575 & \textbf{0.02574} \\
		Loglike   & -3.89 & -3.60 & -1.39   & \textbf{-1.16}   \\ \hline
	\end{tabular}
\end{table}

From these results, we can see that our proposed UTHP model achieves obvious improvement than the baselines among the different scenarios, different data sets have markedly different characteristics, for instance, the number of classes of MIMIC-II dataset is 75, and the average length is only 4, while the number of classes of Financial dataset is only 2, and the average length is 2074. In all of these datasets, our model improves relatively significantly. This suggests that UTHP can better model complex asynchronous event sequences and learn long-term and short-term event dependencies than existing baseline models.

\begin{figure}[!htbp]
	\label{fig3}
	\centering
	\subfigure[MIMIC-II]{
		\includegraphics[width=4.5cm]{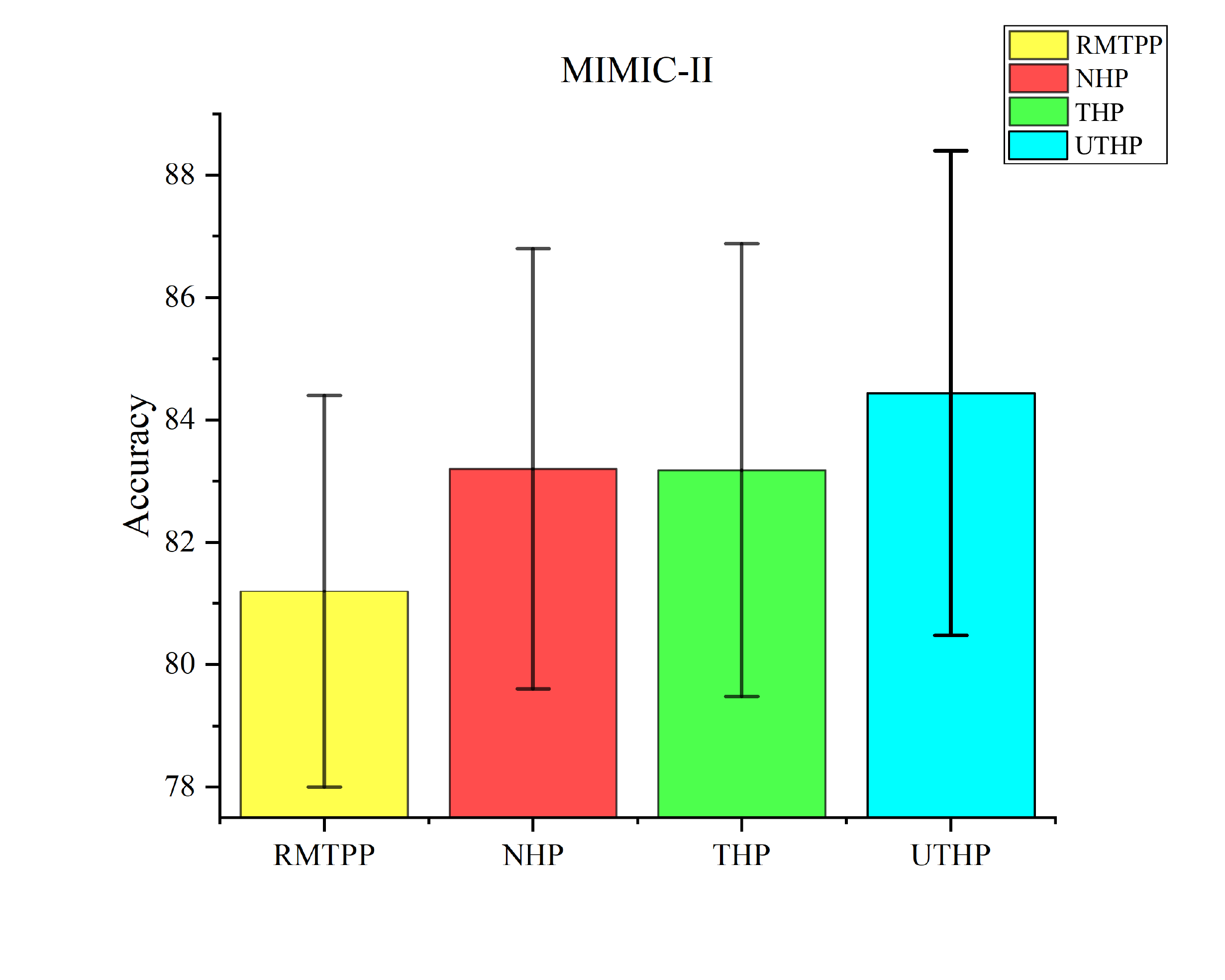}
	}
	\quad
	\subfigure[StackOverflow]{
		\includegraphics[width=4.5cm]{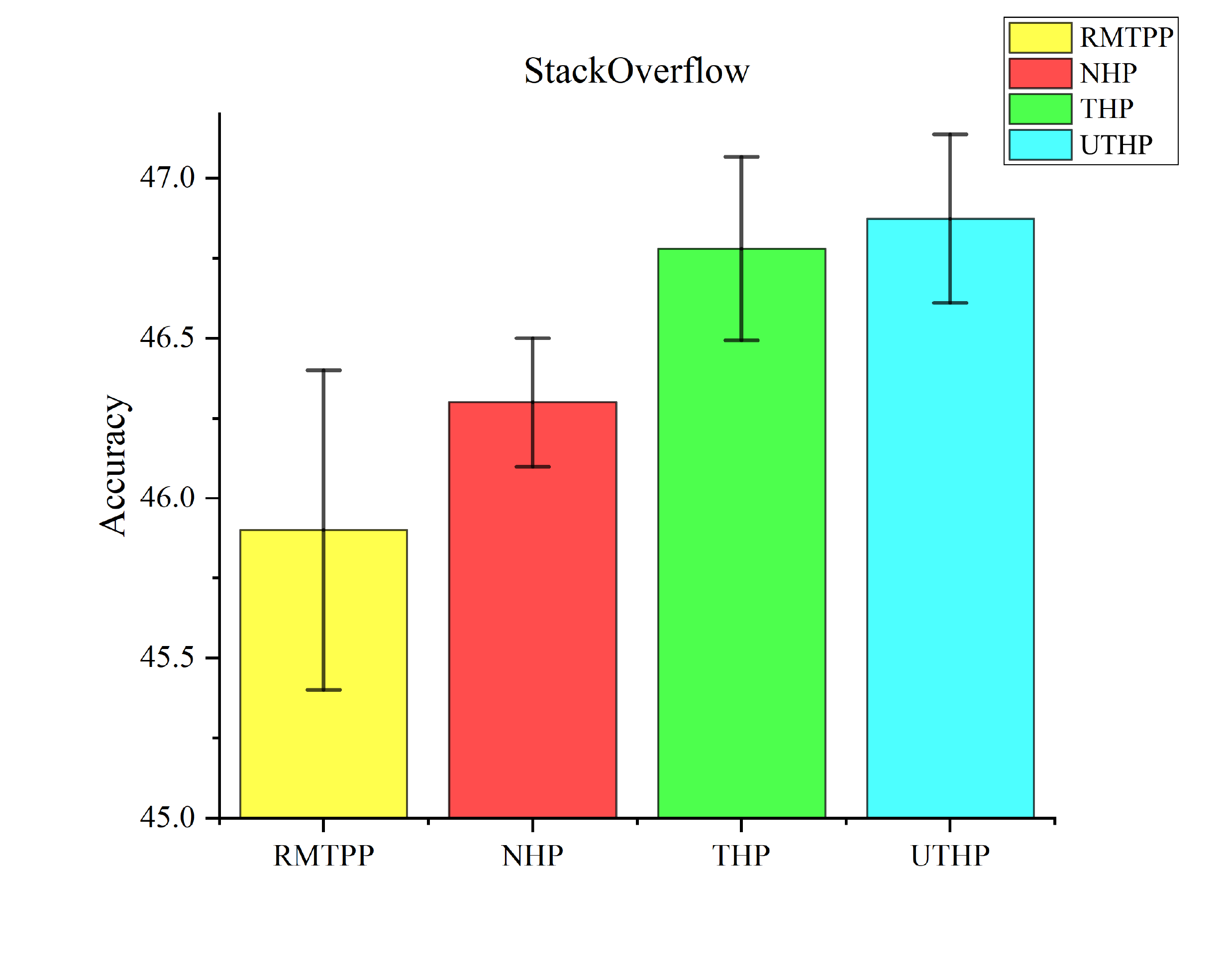}
	}
	\quad
	\subfigure[Fianancial]{
		\includegraphics[width=4.5cm]{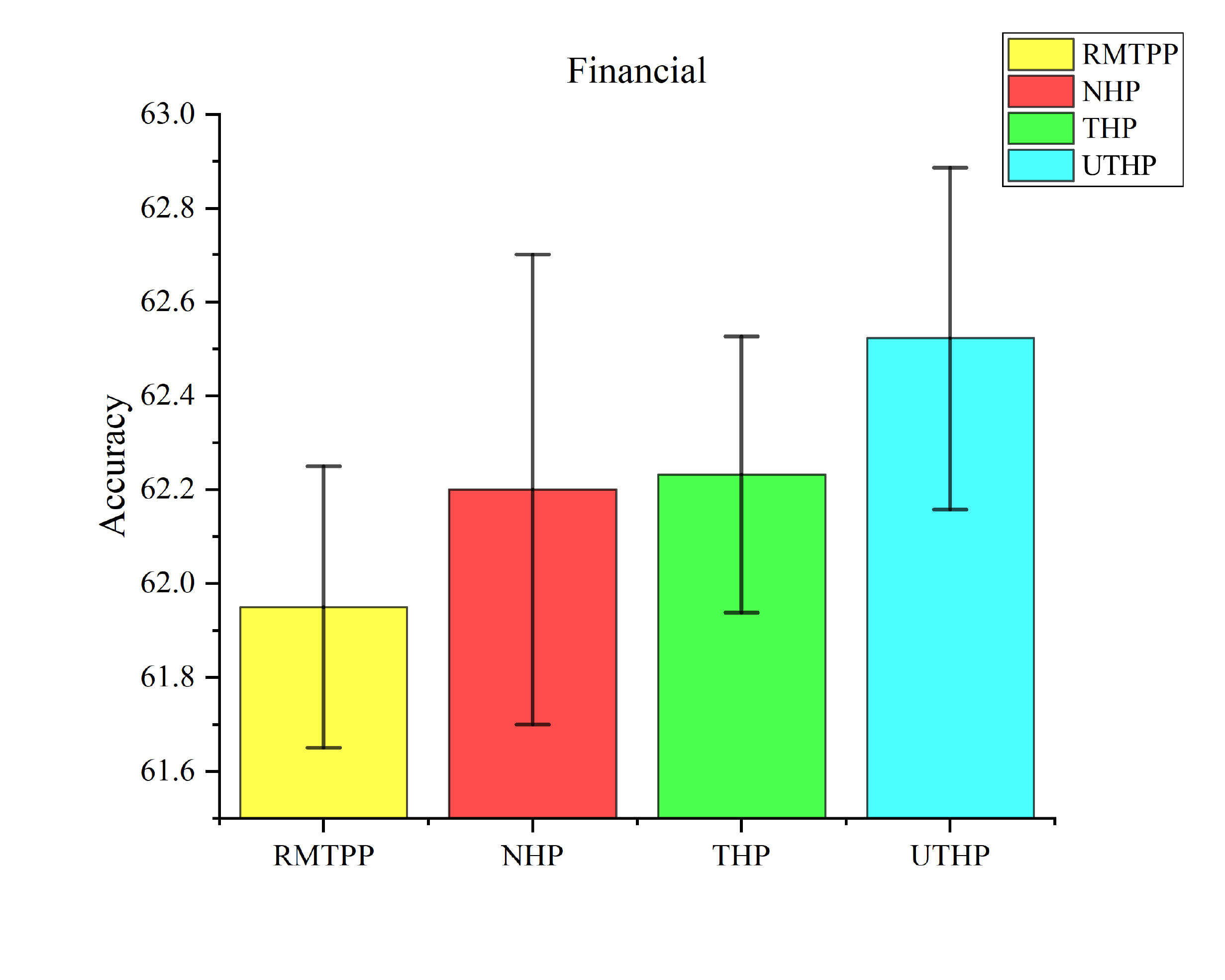}
		%\caption{fig1}
	}
	\caption{Prediction accuracies of RMTPP, NHP, THP and UTHP. Based on the five times train-dev-test partition, five experiments are performed on each dataset, the mean and standard deviations of different models are depicted.}
\end{figure}

\begin{figure}[!htbp]
	\centering
	\subfigure[MIMIC-II]{
		\includegraphics[width=4.5cm]{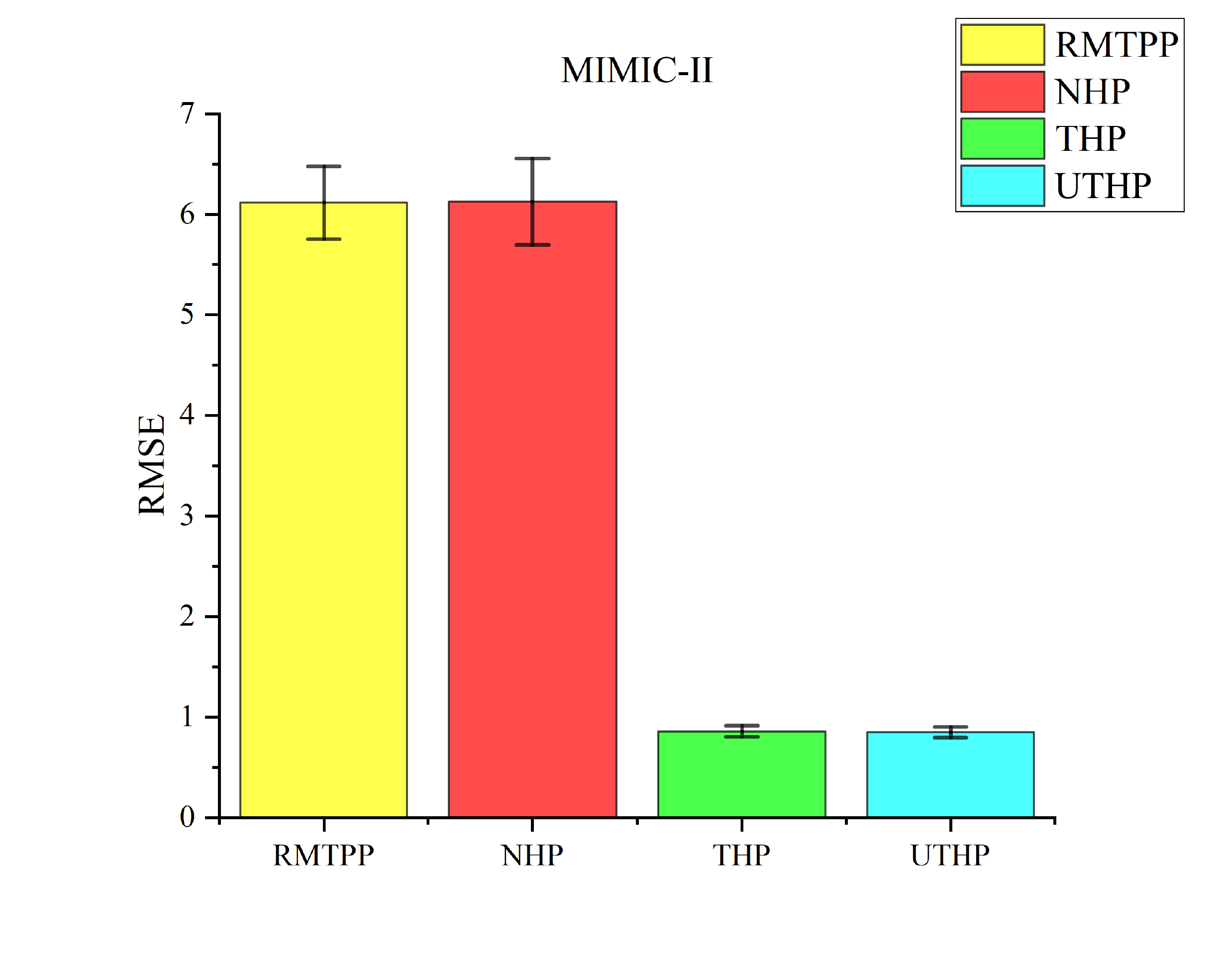}
	}
	\quad
	\subfigure[StackOverflow]{
		\includegraphics[width=4.5cm]{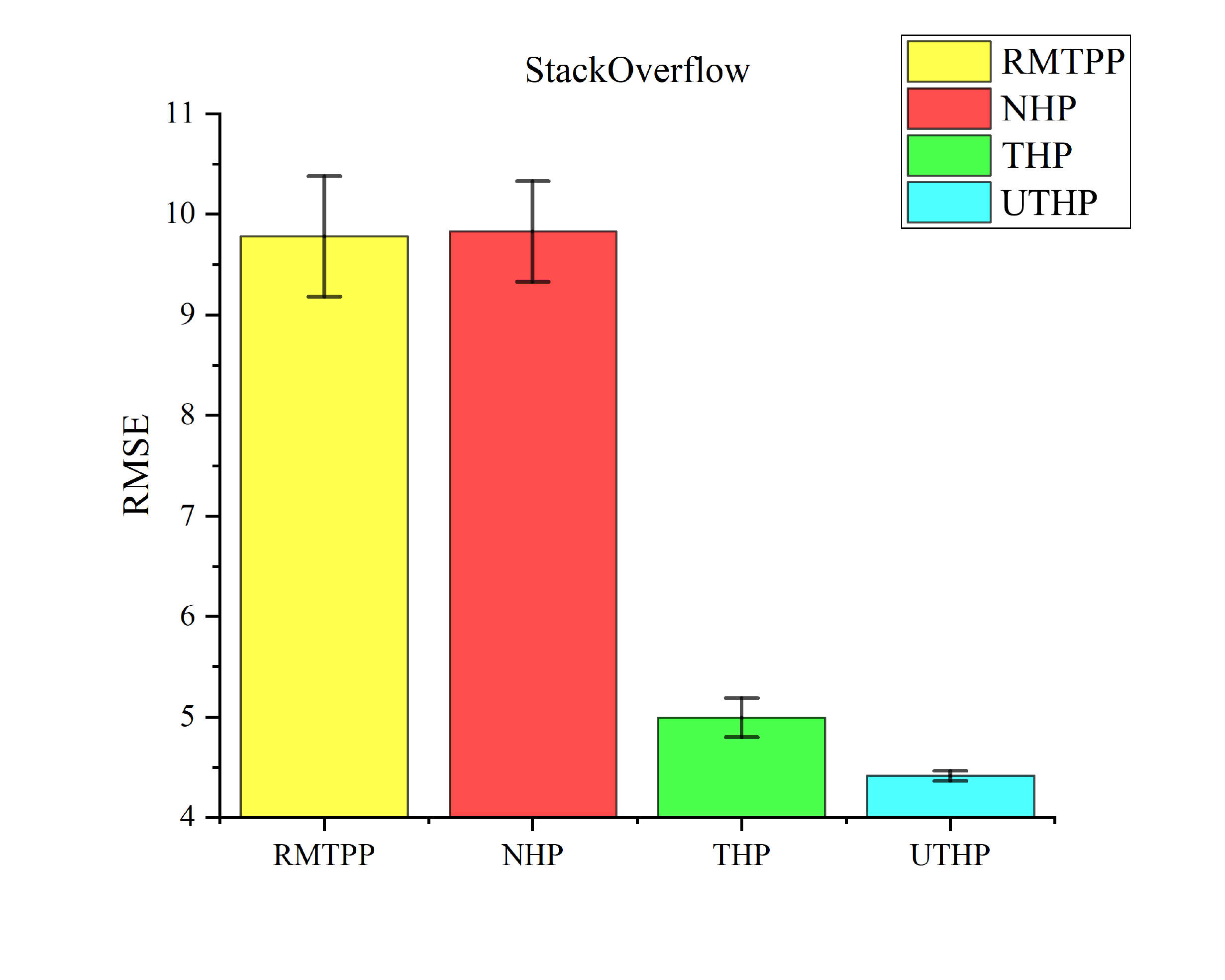}
	}
	\quad
	\subfigure[Fianancial]{
		\includegraphics[width=4.5cm]{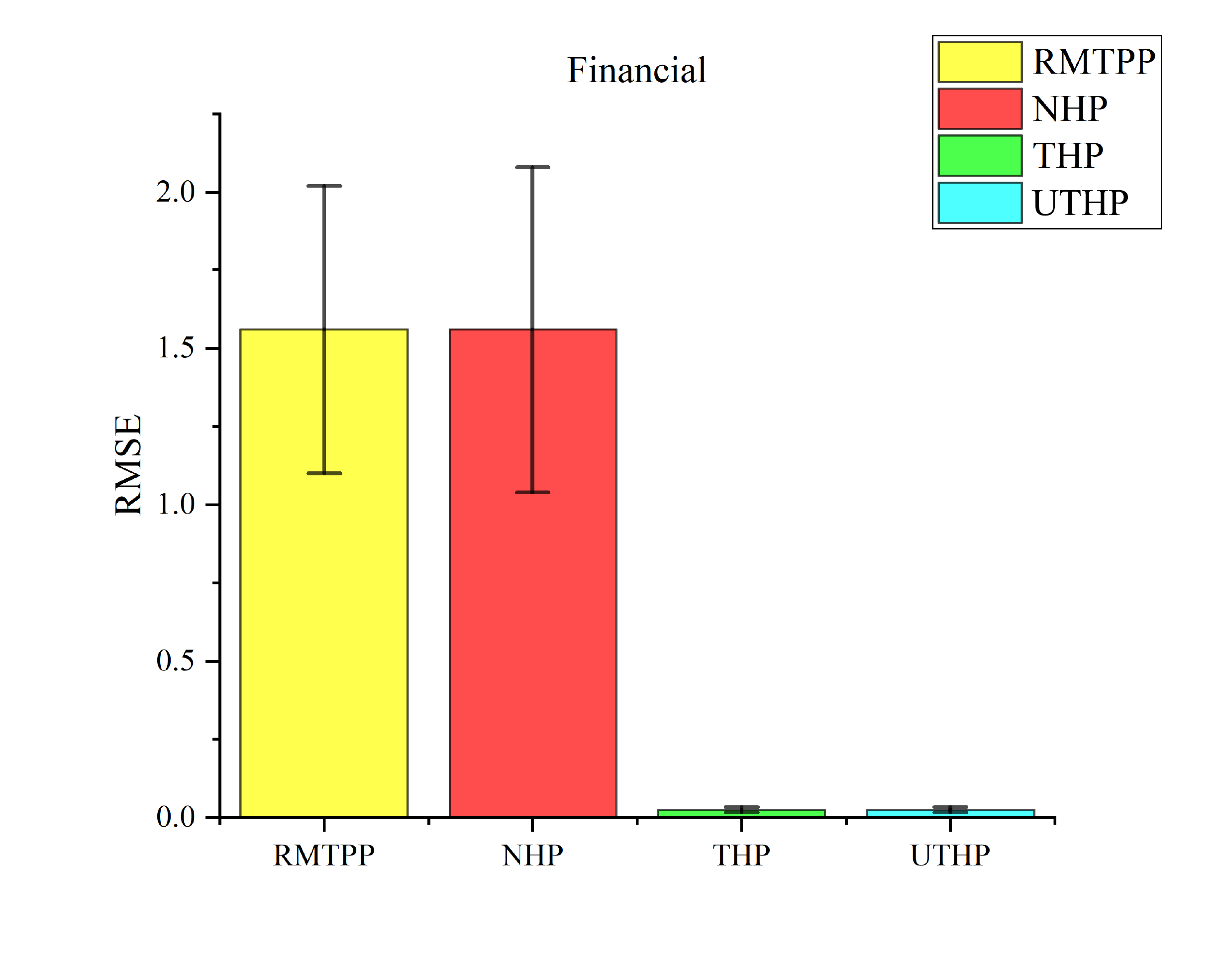}
		%\caption{fig1}
	}
	\subfigure[Zoom in of RMSE of THP and UTHP on the datasets]{
	\includegraphics[width=7cm]{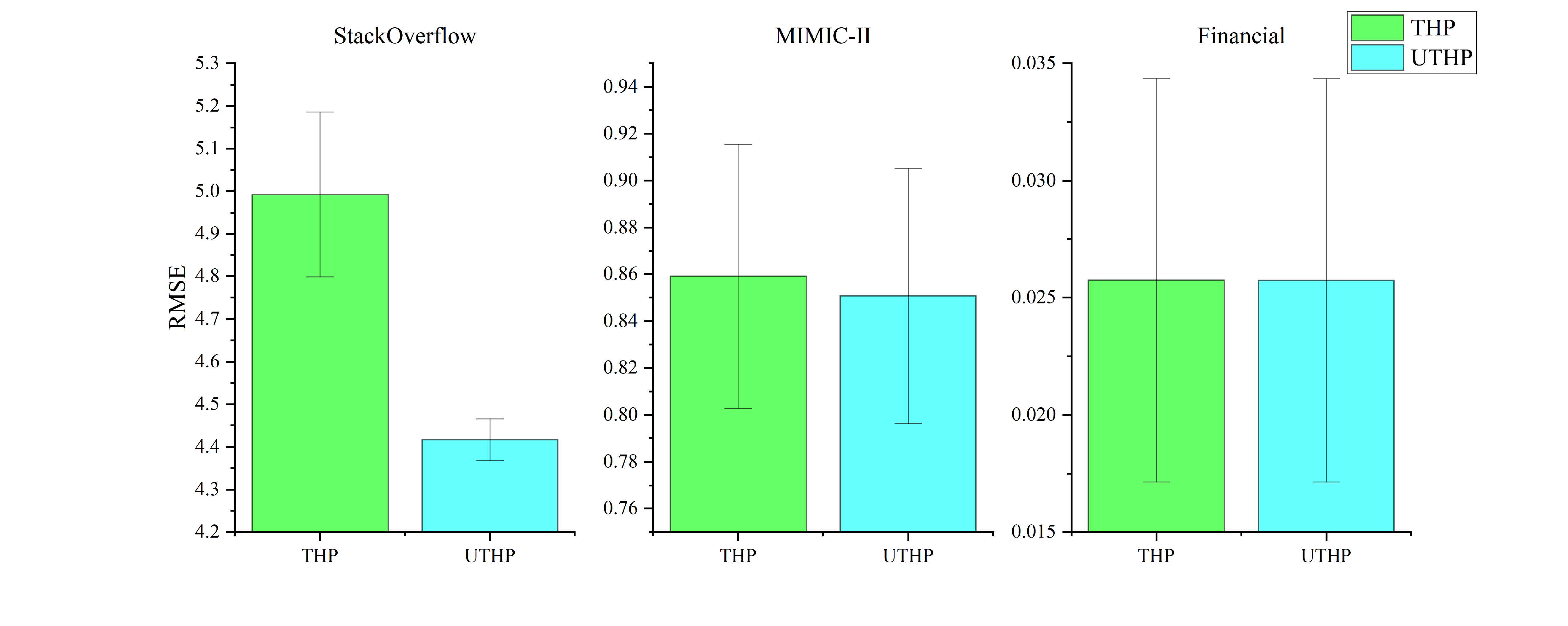}
	%\caption{fig1}
	}
	\caption{The mean and standard deviation for different models obtained from the five experiments in term of RMSE metric for RMTPP, NHP, THP and UTHP.}
\end{figure}

Fig. 5 and Fig.6 visualizes the accuracies of different models and RMSE of baselines and UHP, due to the dispersion of the data, the error bars are wide. We can find out that the results of UTHP better than the other baselines.

\subsection{Ablation study}

In this subsection, we are going to discuss the impact to the performance of three significance part in models, the postprocessing part, CNN module in the position-wise-feed-forward part and ACT mechanism, we demonstrate experimentally how the presence of these three parts affects the performance of the model.

\subsubsection{Ablation study of the postprocessing part}

As we mentioned in subsection 3.1, the introduction of the additional postprocessing part will have an impact on the transformer-based models, and the impact on the model varies with the dataset. In this subsection, we will demonstrate the impact of the introduction of RNN.

Previous experimental results prove that RNN (including LSTM and GRU) can successfully capture the short-term dependencies in the sequence, but not for long-term dependencies. Thus, we can assume that if we introduce the RNN in the additional postprocessing part to the transformer-based model, it will have better performance to capture the dependencies between the events on the dataset with short-term characteristic, and have relatively poor performance of dependencies fitting on the dataset with long-term characteristic.

To validate whether the facts are consistent with our assumptions, validation experiments are carried out on the dataset with long-term property, i.e., MIMIC-II, and dataset with short-term property, i.e., Financial dataset, and the event prediction accuracies are used to measure how well the model fits the dependencies between events. The experimental results are shown as Table 7 and Fig. 7:

\begin{table}[]
	\centering
	\caption{Consider that whether there is an additional RNN layer in the postprocessing part, UTHP and THP’s event prediction accuracy rates on MIMIC-II dataset and financial dataset.}
	\begin{tabular}{ccccc}
		\hline
		& \multicolumn{2}{c}{MIMIC-II} & \multicolumn{2}{c}{Financial} \\ \hline
		& With RNN    & Without RNN    & With RNN     & Without RNN    \\ \hline
		\begin{tabular}[c]{@{}c@{}}Accuracy of THP\end{tabular}  & 81.44       & \textbf{83.18}          &\textbf{ 62.23}        & 60.22          \\
		\begin{tabular}[c]{@{}c@{}}Accuracy of UTHP\end{tabular} & 83.34       & \textbf{84.43 }         & \textbf{62.52 }       & 62.04          \\ \hline
	\end{tabular}
\end{table}

From the experimental results, we can discover that for transformer-based model, addition RNN layer in the postprocessing part has a negative effect on the experimental results of MIMIC-II dataset, and has a positive impact on the financial dataset. These facts verify that our hypothesis, RNN do enhance the fitting ability to events short-term dependencies, and weaken the fitting ability to the events long-term dependencies.

\begin{figure}[!htbp]
	\label{fig3}
	\centering
	\subfigure[MIMIC-II]{
		\includegraphics[width=4.5cm]{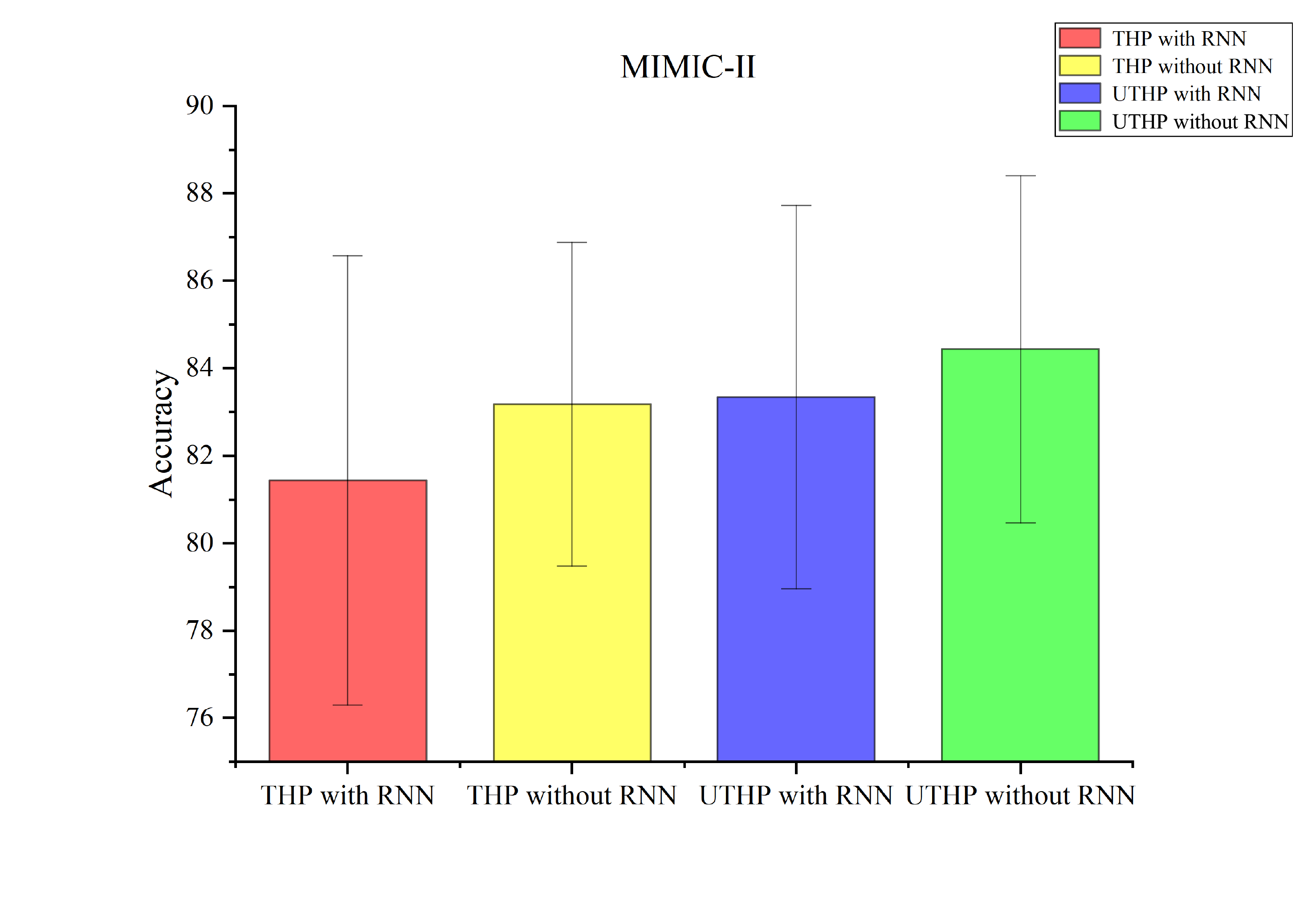}
	}
	\quad
	\subfigure[Financial]{
		\includegraphics[width=4.5cm]{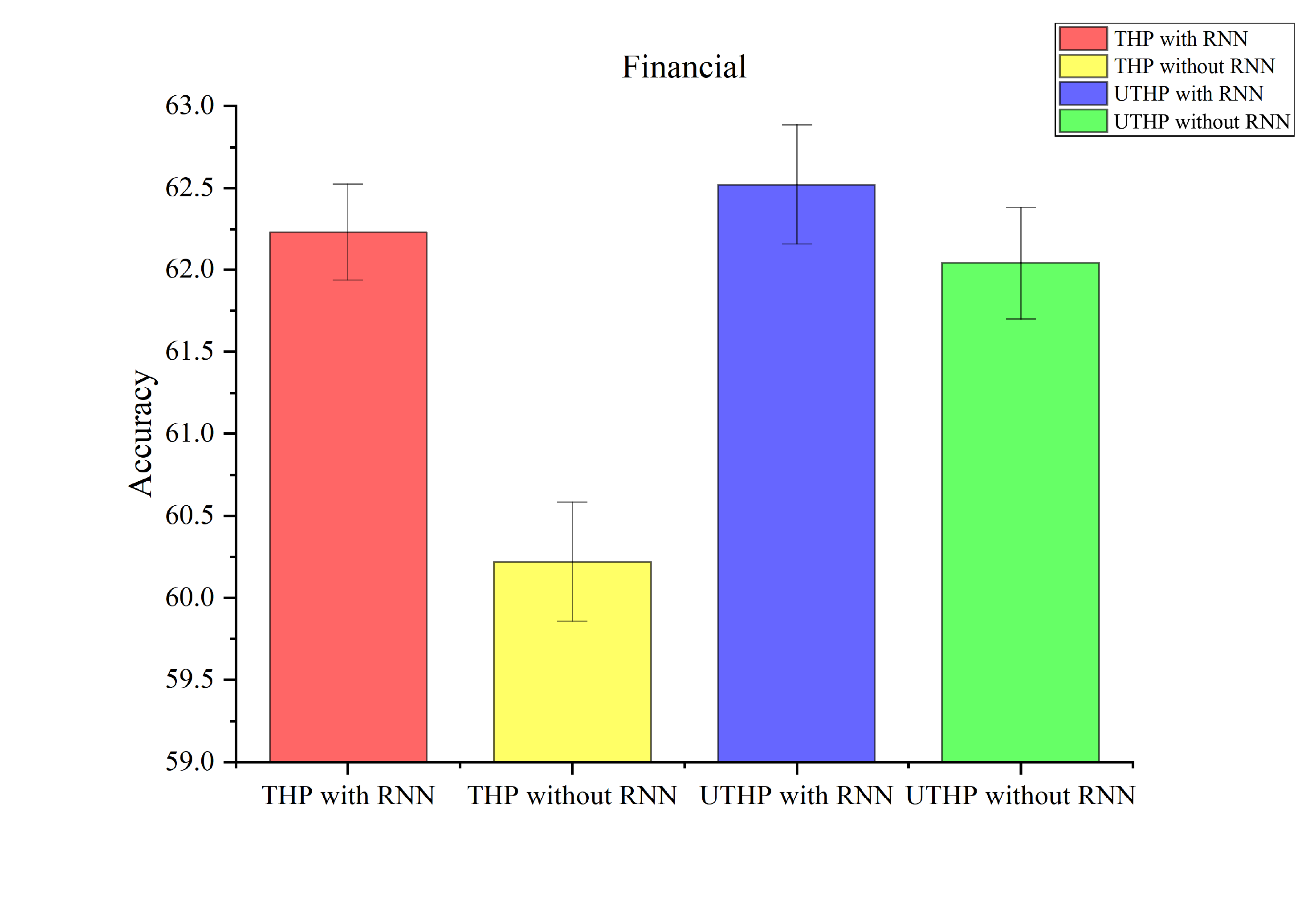}
	}
	\quad
	\caption{Visualization of ablation experimental results with and without an additional RNN layer in the postprocessing part.}
\end{figure}

In addition, UTHP model demonstrates its robustness without RNN on the financial dataset. The reason is that our model already has a recursive structure, which helps it better model short-term data. This also reflects the value of UTHP model we proposed.

\subsubsection{Ablation study of CNN module in the position-wise-feed-forward part}

In general, CNN module will enhance the local perception ability of the model, which is the reason why we add CNN module into the position-wise-feed-forward part. In order to judge whether the introduction of the CNN module is effective, we conduct the comparison experiments, compare the UTHP based on ACT algorithm with CNN and without CNN module in position-wise-feed-forward part on StackOverflow, MIMIC-II and Financial dataset. The experiment results are shown as Table 8 and Fig. 8.

\begin{table}[]
	\centering
	\caption{Model performance comparisons of UTHP based on ACT algorithm with CNN and without CNN module in position-wise-feed-forward part}
	\begin{tabular}{cccc}
		\hline
		& UTHP          & With   CNN & Without   CNN \\ \hline
		\multirow{3}{*}{Accuracy} & StackOverflow & \textbf{46.87}      & 46.76         \\
		& MIMIC-II      & \textbf{84.43}      & 83.83         \\
		& Financial     & 62.52      & \textbf{62.52}         \\ \hline
	\end{tabular}
\end{table}

\begin{figure}[!htbp]
	\centering
	\includegraphics[scale=0.3]{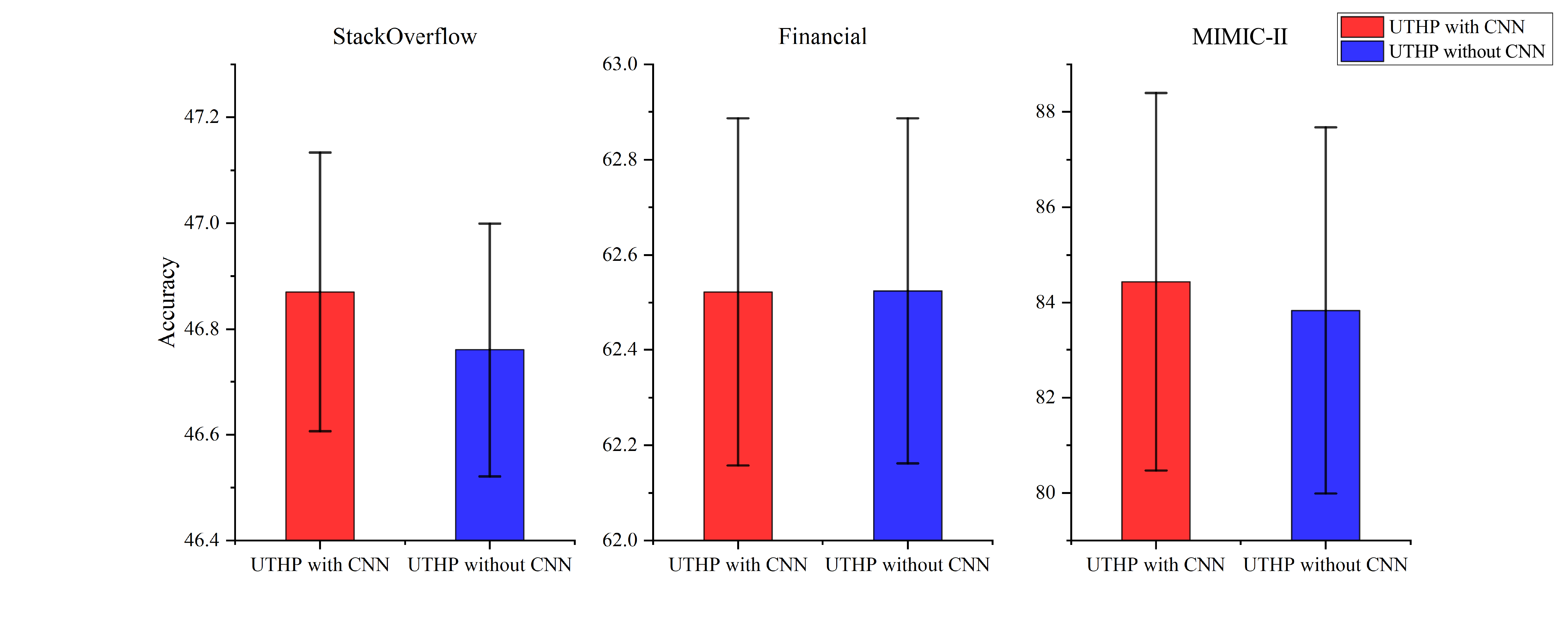}
	\caption{Accuracy comparison of UTHP based on ACT algorithm with CNN and without CNN module in position-wise-feed-forward part.}
	\label{fig3}
\end{figure}

\subsubsection{Ablation study of ACT mechanism}

From the above experimental results, we can find out that the introduction of CNN module in position-wise-feed-forward part obviously raises the event prediction accuracy. This fact indicates that CNN module enhance the ability to model connections between events of UTHP through its local perception characteristics, which indicates that the introduction of CNN module in position-wise-feed-forward part is effective.

In subsection 3.1, we introduce the reasons for adding the ACT mechanism to the model, ACT mechanism will dynamically regulate the calculation of input in the transformer, make the model use more computing resources to where it is needed more. The concept of this effect is similar to and complementary to the self-attention mechanism, and theoretically makes the model perform better.
And in order to investigate the effect of the ACT mechanism on UTHP, we conducted a series of comparative experiments, we compare pure UTHP under different recursive iteration times (the corresponding algorithm is Algorithm 1) and UTHP with ACT mechanism (the corresponding algorithm is Algorithm 2 and maximum number of iterations max\_n=2). These results are summarized in Table 9:

\begin{table}[]
	\centering
	\caption{Model performance comparisons of UTHP based on ACT mechanism and UTHP with pure recurrence}
	\begin{tabular}{ccccccc}
		\hline
		& Times of iteration & ACT     & 1       & 2       & 3       & 4       \\\hline
		\multirow{3}{*}{StackOverflow} & Accuracy           & \textbf{46.87}   & 46.67   & 46.70   & 46.70   & 46.72   \\
		& RMSE               & \textbf{4.42}    & 4.52    & 4.47    & 4.47    & 4.47    \\
		& Loglike            & \textbf{-0.55}   & -0.60   & -0.58   & -0.58   & -0.57   \\
		\multirow{3}{*}{MIMIC-II}      & Accuracy           & \textbf{84.43}   & 82.34   & 82.63   & 83.05   & 81.95   \\
		& RMSE               & \textbf{0.85}    & 0.90    & 0.87    & 0.89    & 0.85    \\
		& Loglike            & -0.18   & -0.21   & -0.19   & -0.18   & \textbf{-0.17}   \\
		\multirow{3}{*}{Financial}     & Accuracy           & \textbf{62.52}   & 62.52   & 62.45   & 62.43   & 62.29   \\
		& RMSE               & 0.02574 & 0.02574 & 0.02572 & \textbf{0.02571} & 0.02572 \\
		& Loglike            & -1.16   & -1.24   & \textbf{-0.76}   & -0.80   & -0.77   \\ \hline
	\end{tabular}
\end{table}

For the pure recurrence of UTHP, we considered four different iteration times as the first row in Table 9, and when the number of iterations is once, it indicates that hidden representation of sequences will directly output the model, rather than recurrently iterate in the model, from the results we can find out that UTHP with ACT mechanism performs than pure recurrence UTHP overall. Especially in terms of the accuracies of event prediction, UTHP with ACT mechanism are all better than UTHP with pure recurrence.

For a more intuitive comparison, we visualize these experimental results and error bars, as shown in Fig. 9, Fig. 10, and Fig. 11.

\begin{figure}[!htbp]
	\label{fig3}
	\centering
	\subfigure[Accuracy]{
		\includegraphics[width=4.5cm]{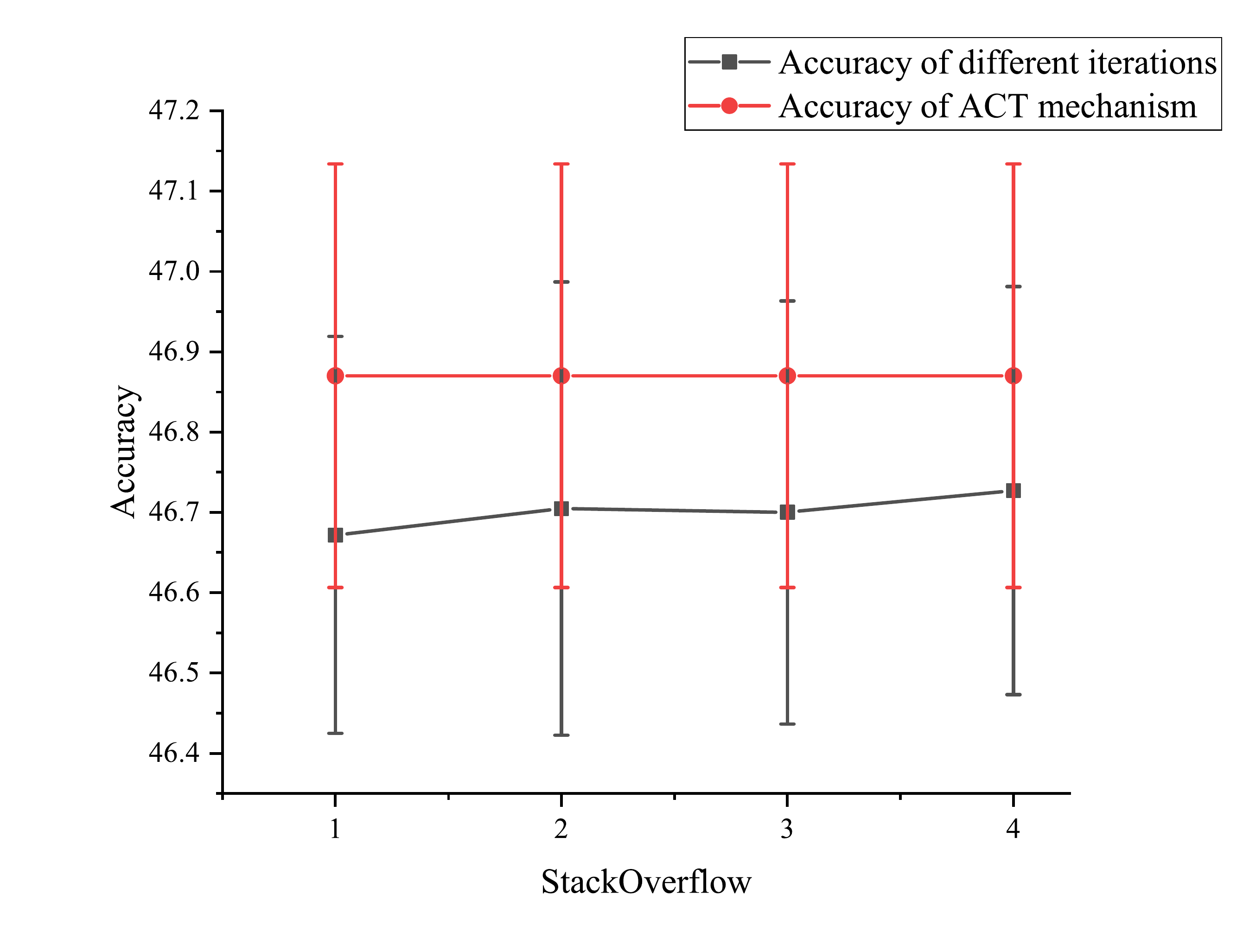}
	}
	\quad
	\subfigure[RMSE]{
		\includegraphics[width=4.5cm]{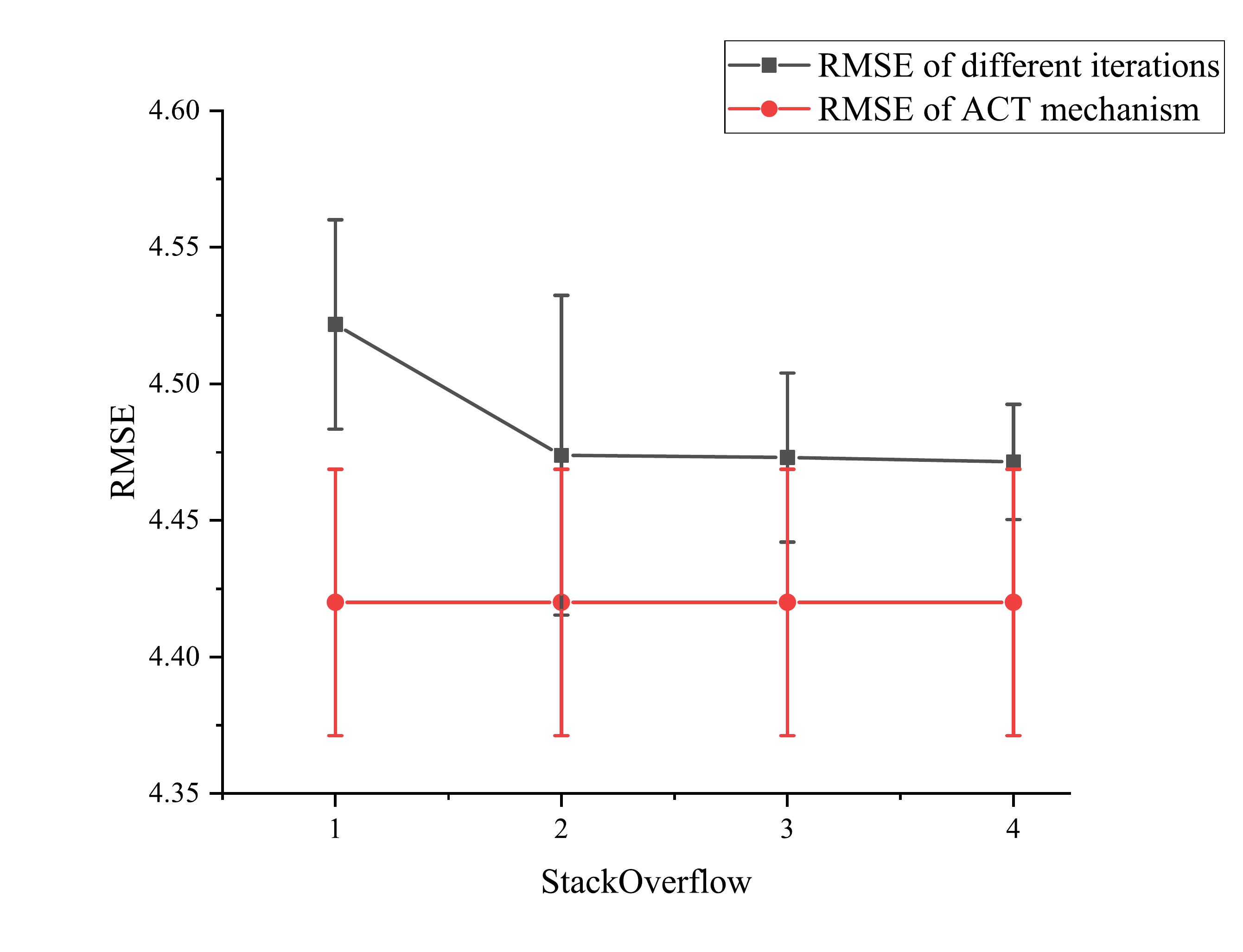}
	}
	\quad
	\subfigure[Loglike]{
	\includegraphics[width=4.5cm]{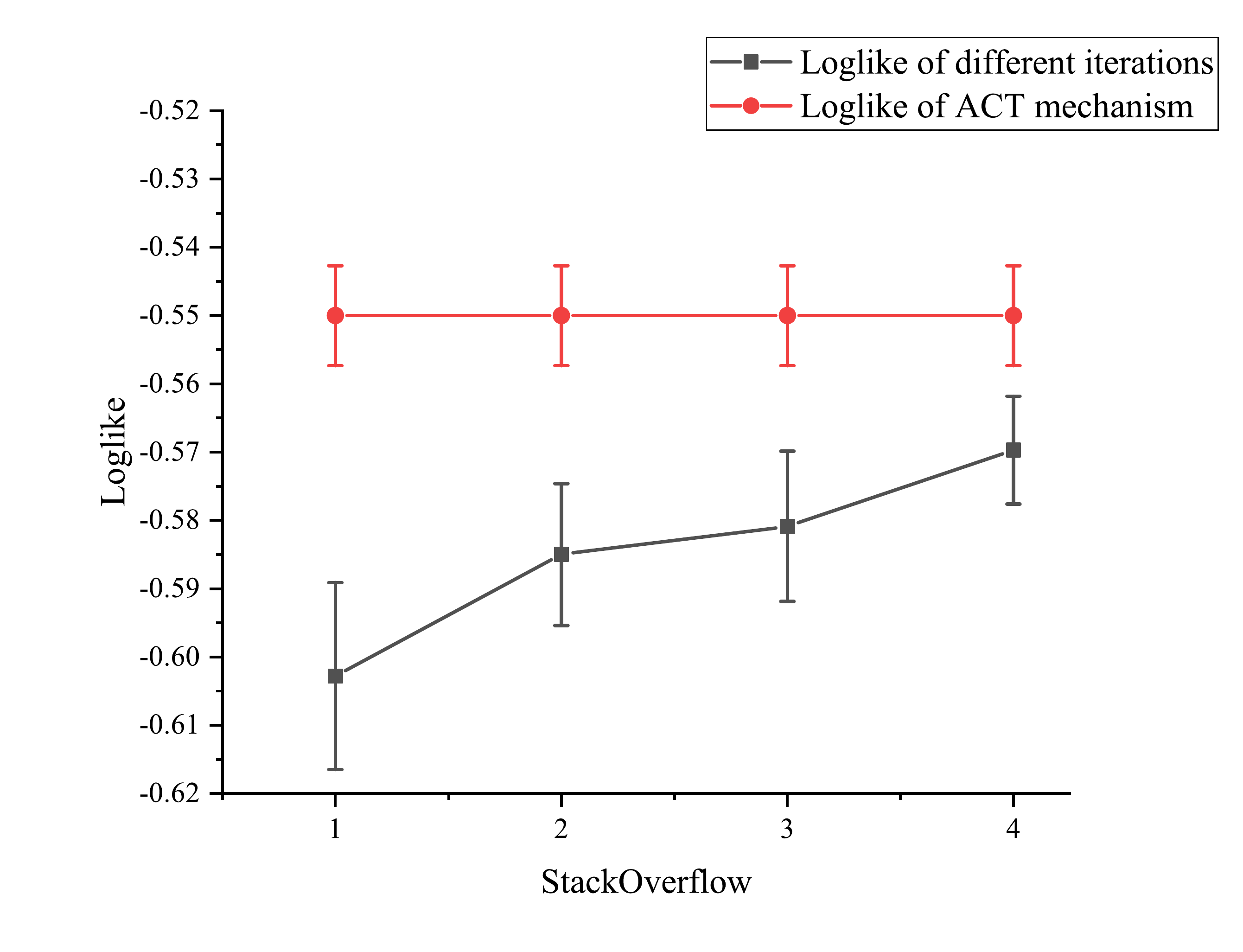}
	}
	\caption{Model performance curves of UTHP with ACT mechanism and pure UTHP on StackOverflow dataset.}
\end{figure}

\begin{figure}[!htbp]
	\label{fig3}
	\centering
	\subfigure[Accuracy]{
		\includegraphics[width=4.5cm]{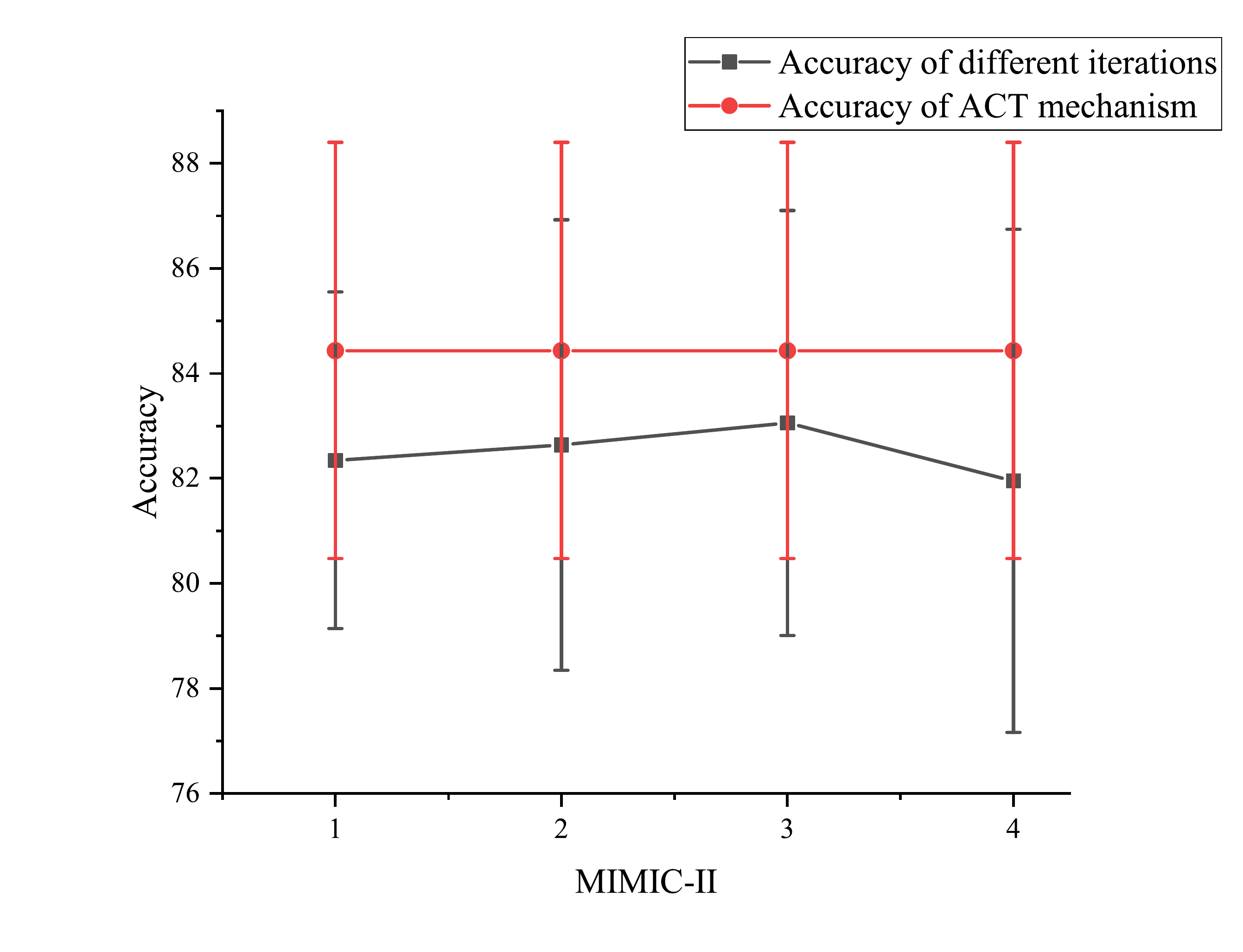}
	}
	\quad
	\subfigure[RMSE]{
		\includegraphics[width=4.5cm]{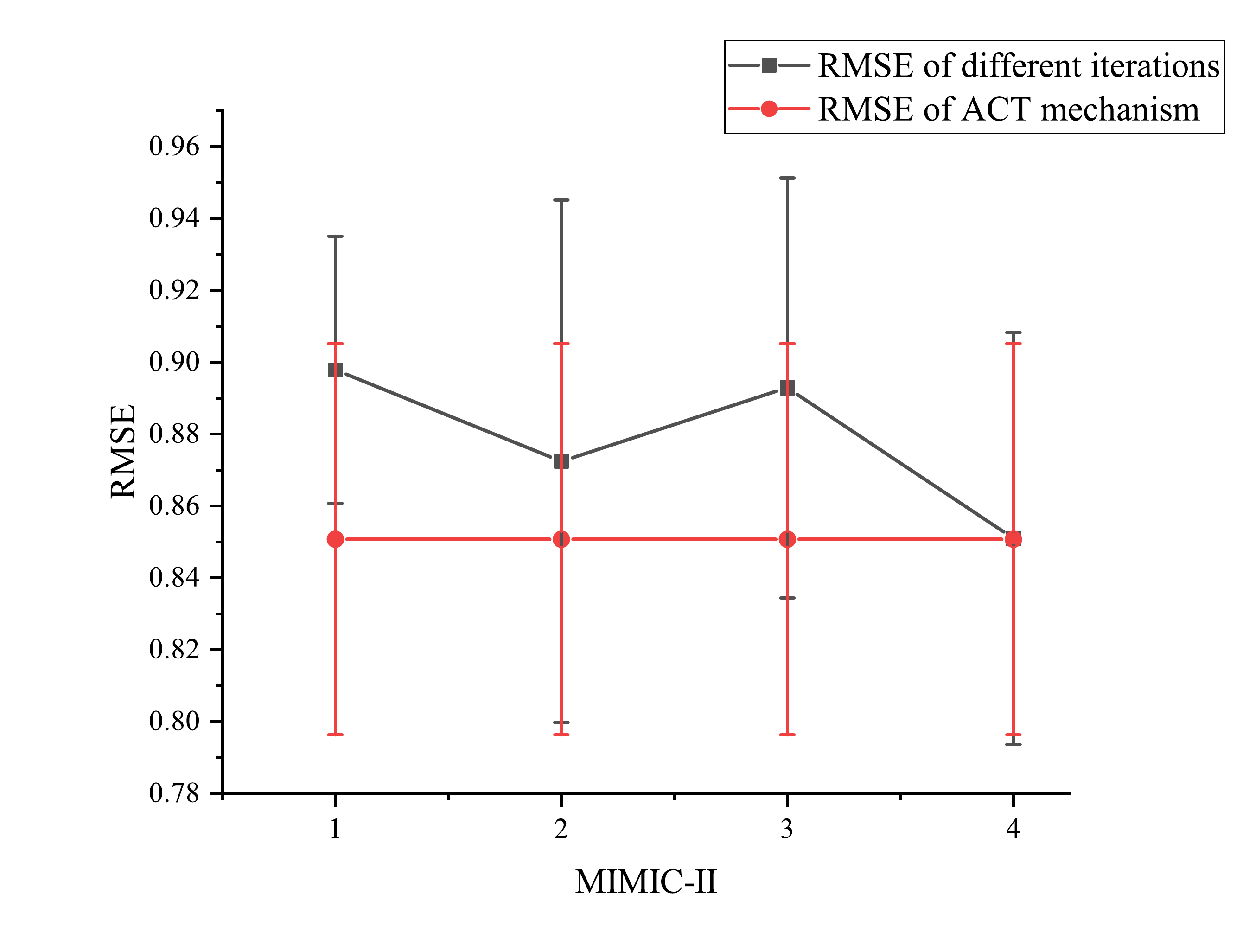}
	}
	\quad
	\subfigure[Loglike]{
		\includegraphics[width=4.5cm]{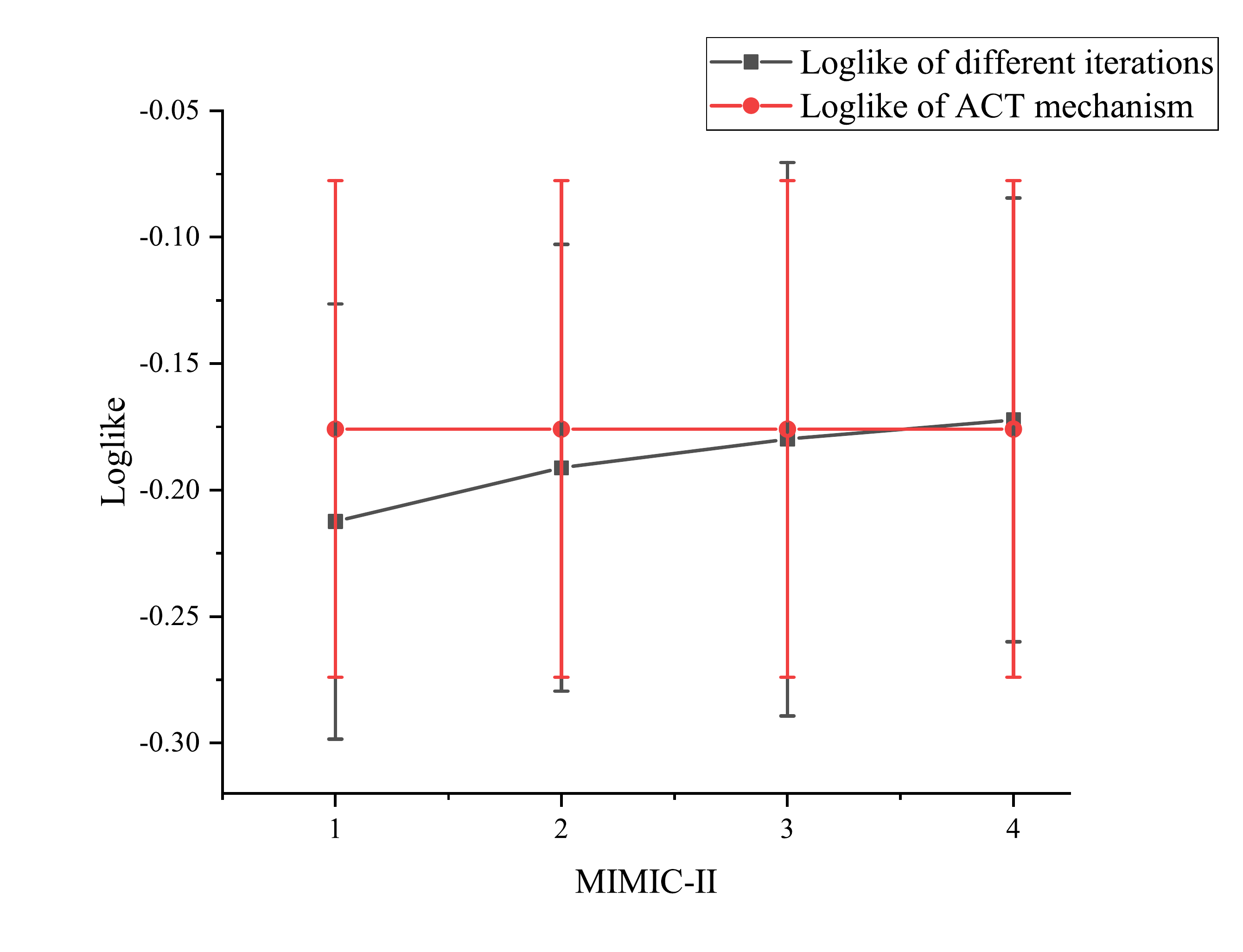}
	}
	\caption{Model performance curves of UTHP with ACT mechanism and pure UTHP on MIMIC-II dataset.}
\end{figure}

\begin{figure}[!htbp]
	\label{fig3}
	\centering
	\subfigure[Accuracy]{
		\includegraphics[width=4.5cm]{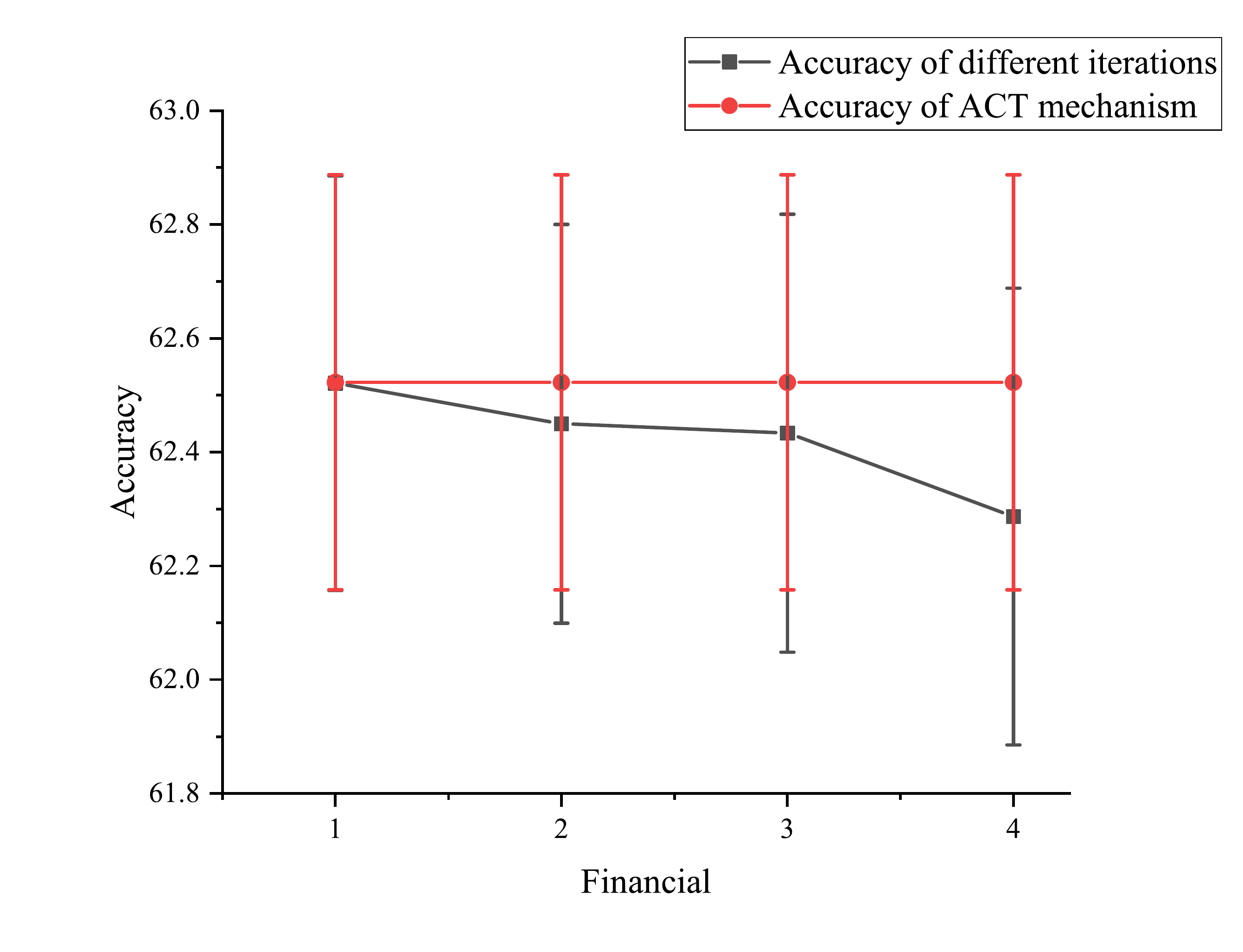}
	}
	\quad
	\subfigure[RMSE]{
		\includegraphics[width=4.5cm]{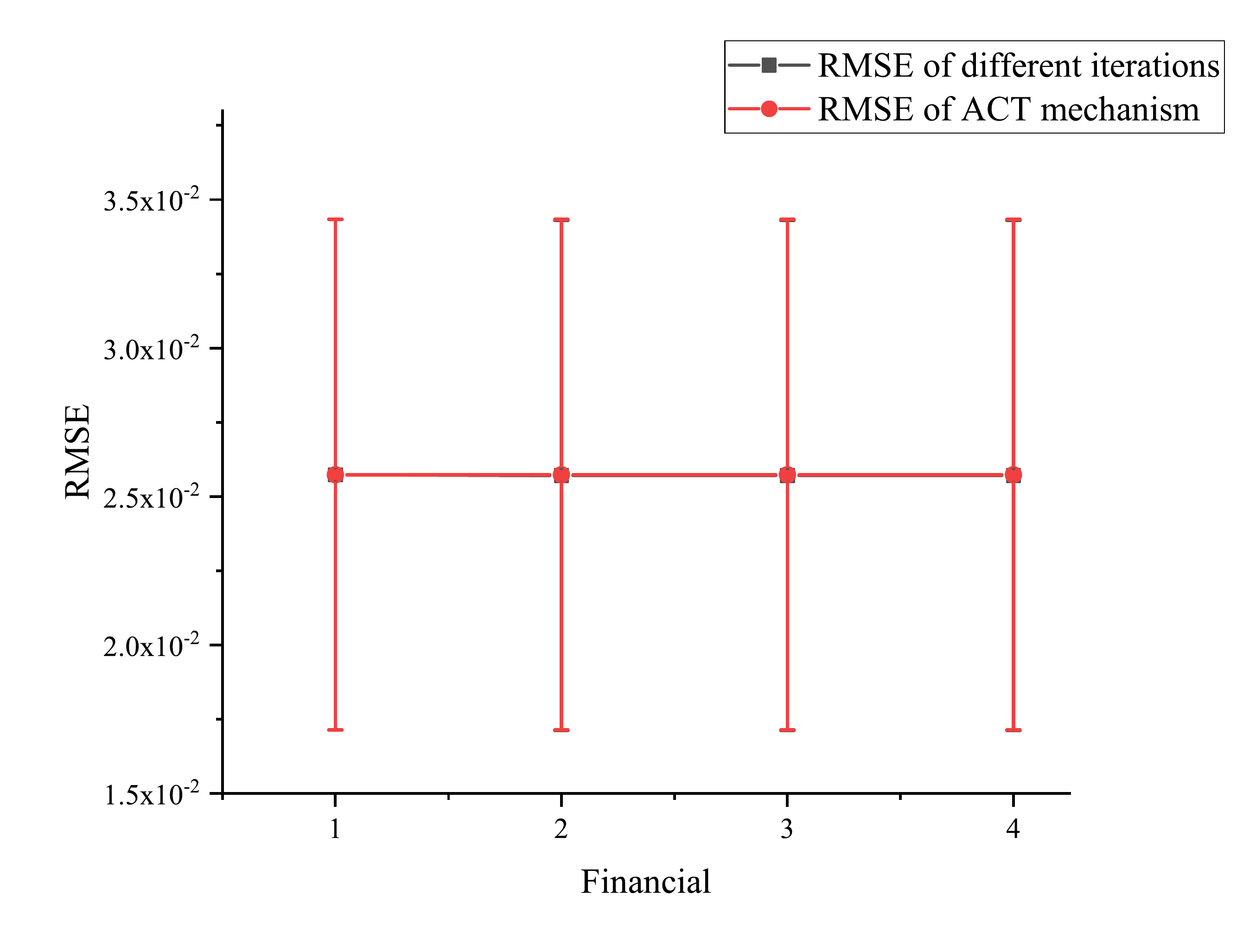}
	}
	\quad
	\subfigure[Loglike]{
		\includegraphics[width=4.5cm]{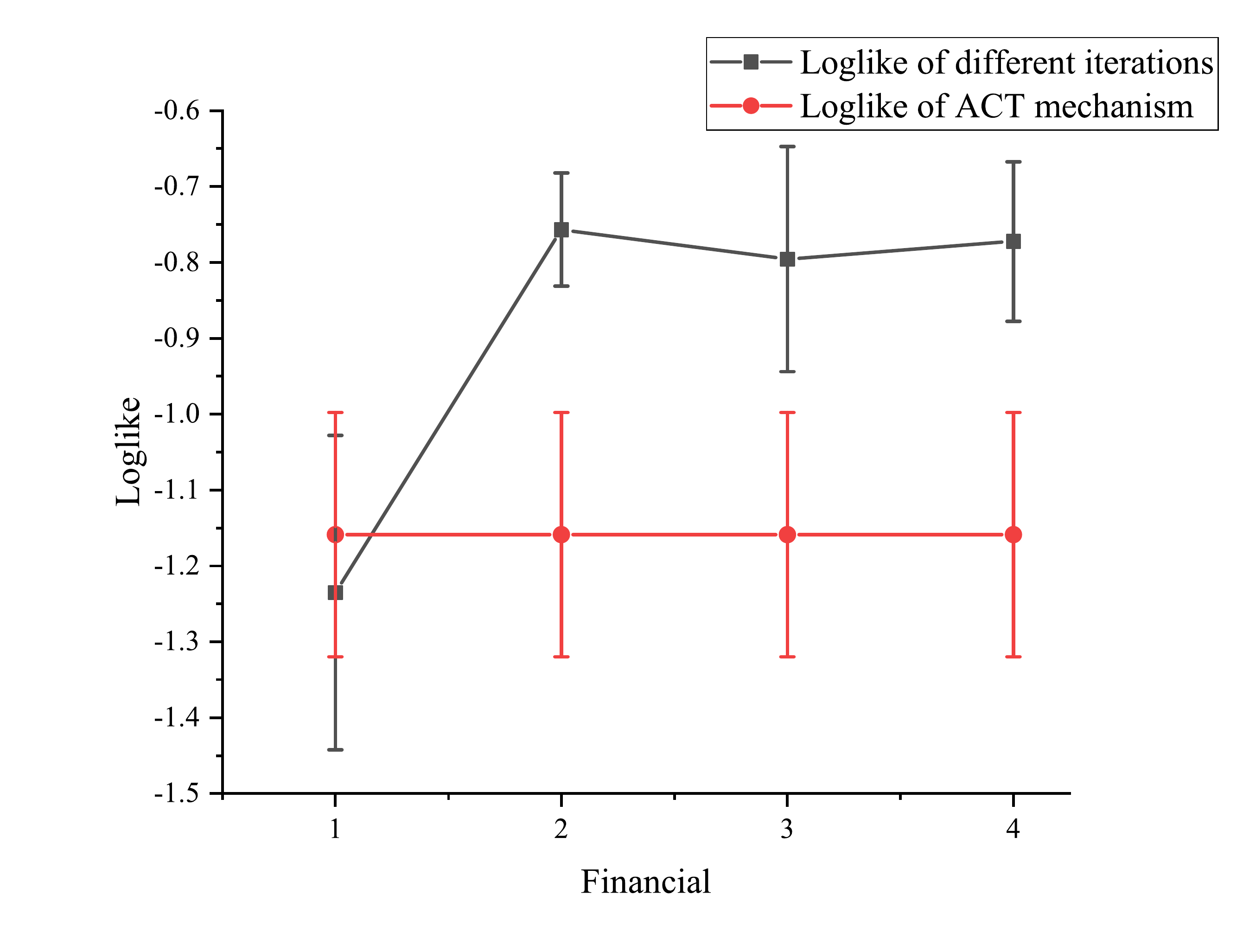}
	}
	\caption{Model performance curves of UTHP with ACT mechanism and pure UTHP on Financial dataset.}
\end{figure}

From these figures, we can see that compared to the error bar, UTHP with ACT mechanism has a clear gap with pure UTHP when it performs well, such as the accuracy, RMSE and Loglike curves on StackOverflow dataset, and the accuracy curves on MIMIC-II dataset. And comparing with the standard deviation, when pure UTHP results are better than UTHP with mechanism, the difference is not so obvious. This further verifies that the introduction of the ACT mechanism will improve the overall performance of UTHP.

In addition, with the increasing of iteration times, the Loglike on these datasets of pure UTHP are also increasing. The fact represents that the increment of iteration items enhance the modeling ability of UTHP.

\subsection{Fewer model parameters of UTHP}

Because of the recurrence structure of UTHP, there is only one encoding layer in our model, while THP utilize the stacking of multiple encoding layer, in other words, UTHP model share the parameters in the encoding layer. By this way, we find that our proposed architectures improve model performance, and also reduce model parameters.

For instance, in subsection 4.3, on each dataset, UTHP and THP adopt the similar hyper-parameters, which is shown as Table 10 and Table 11, the only difference is that UTHP has only one encoding layer, and there is a CNN module in the position-wise-feed-forward part of encoding layer. 

\begin{table}[]
	\centering
	\caption{Hyperparameter configurations of THP.}
	\begin{tabular}{ccccccc}\hline
		Dataset & Synthetic & Retweets & MemeTrack & Financial & MIMIC-II & \begin{tabular}[c]{@{}c@{}}StackOverflow\end{tabular} \\ \hline
	$D$	& 64        & 64       & 64        & 128       & 64       & 512                                                            \\
	$D_H$	& 256       & 256      & 256       & 2048      & 256      & 1024                                                           \\
	$D_{\rm{RNN}}$	& 128       & 128      & 128       & 128       & 0        & 128                                                            \\
	$D_Q=D_V$	& 16        & 16       & 16        & 64        & 16       & 512                                                            \\
	Heads of attention	& 3         & 3        & 3         & 6         & 3        & 4                                                              \\
	Layers of transformer	& 2         & 2        & 2         & 2         & 2        & 2                                                              \\
	Dropout	& 0.1       & 0.1      & 0.1       & 0.1       & 0.1      & 0.1                                                            \\    \hline                           
	\end{tabular}
\end{table}

\begin{table}[]
	\centering
	\caption{Hyperparameter configurations of UTHP.}
	\begin{tabular}{ccccccc}\hline
		Dataset & Synthetic & Retweets & MemeTrack & Financial & MIMIC-II & \begin{tabular}[c]{@{}c@{}}StackOverflow\end{tabular} \\\hline
	$D$	& 64        & 64       & 64        & 128       & 64       & 512                                                            \\
	$D_H$	& 256       & 256      & 256       & 2048      & 256      & 1024                                                           \\
	$D_{\rm{RNN}}$	& 128       & 128      & 128       & 128       & 0        & 128                                                            \\
	$D_Q=D_V$	& 16        & 16       & 16        & 64        & 16       & 512                                                            \\
	Heads of attention	& 3         & 3        & 3         & 6         & 3        & 4                                                              \\
	Max\_n	& 2         & 2        & 2         & 2         & 2        & 2                                                              \\
	Dropout	& 0.1       & 0.1      & 0.1       & 0.1       & 0.1      & 0.1                                                            \\
	Size of convolution kernel	& 3         & 3        & 3         & 3         & 3        & 3                                                              \\
	Stride of convolution kernel	& 2         & 2        & 2         & 2         & 2        & 2                                                              \\
	Padding	& 0         & 0        & 0         & 0         & 0        & 0                                                              \\
	Stride of the pooling module	& 2         & 2        & 2         & 2         & 2        & 2                                                              \\
	Size of the pooling module	& 2         & 2        & 2         & 2         & 2        & 2\\ \hline                                                            
	\end{tabular}
\end{table}

\begin{table}[]
	\centering
	\caption{The numbers of parameters of the THP and UTHP on different datasets.}
	\begin{tabular}{ccc}\hline
		Number of parameters & THP      & UTHP    \\\hline
		Synthetic            & 245765   & 129494  \\
		Retweet              & 137795   & 46356   \\
		MemeTrack            & 1102216  & 1010777 \\
		MIMIC-II             & 151691   & 60252   \\
		StackOverflow        & 21022742 & 5281319 \\
		Financial            & 4343298  & 724627 \\\hline
	\end{tabular}
\end{table}

As we can see from the Table 12, although we introduce a CNN module in position-wise-feed-forward part which increases the additional parameters in UTHP, it still has less parameter than THP. The parameter number of the THP’s parameter is approximately 6 to 1.1 times of the UTHP’s one. The fewer number of event types in the dataset, the fewer parameters for UTHP compared to THP. Combined with the experimental results in subsection 4.3, our model uses fewer parameters to achieve better performance, which reflects the improvement of our model.

\section{Conclusions and future works}
In this paper, we come up with UTHP, a new neural point process model to analyze the asynchronous event sequence. UTHP combines the self-attention mechanism in transformer and the recurrence mechanism in RNN, this operation allows our model to organically integrate the advantages of RNN and transformer, moreover, in order to make UTHP to adaptively determine when to stop refining processes of hidden variables, we introduce the ACT mechanism to UTHP. Experimental results verify that our model performs better than the baselines overall with fewer model parameters, and we also discuss the impact of the existence of the additional RNN layer and the ACT mechanism in the ablation study.

\section{Acknowledgment}
 Thanks for Hong-yuan Mei and Si-miao Zuo for their generous help in my research state, their help greatly improved our research.

\bibliography{mybibfile}

\end{document}